\documentclass[acmtog]{acmart}
\acmSubmissionID{517}

\usepackage{flushend}
\usepackage{booktabs} % For formal tables
\usepackage[ruled]{algorithm2e} % For algorithms
\usepackage{algorithmic}
\usepackage{makecell}
\usepackage{multirow}
\usepackage{rotating}
\usepackage{color}

\SetAlFnt{\small}
\SetAlCapFnt{\small}
\SetAlCapNameFnt{\small}
\SetAlCapHSkip{0pt}

\begin{document}
% Title portion
\title{Low-Light Image Enhancement with Wavelet-based Diffusion Models}
\author{Hai Jiang}
\affiliation{
	\institution{Sichuan University, Megvii Technology}
	  \country{China}}
  \email{jianghai1@stu.scu.edu.cn}
  
  \author{Ao Luo}
  \affiliation{
  	\institution{Megvii Technology}
  	\country{China}}
  \email{luoao02@megvii.com}
  
    \author{Songchen Han}
  \affiliation{
  	\institution{Sichuan University}
  	\country{China}}
  \email{hansongchen@scu.edu.cn}
  
      \author{Haoqiang Fan}
  \affiliation{
  	\institution{Megvii Technology}
  	\country{China}}
  \email{fhq@megvii.com}
  
        \author{Shuaicheng Liu}\authornote{Corresponding author}
  \affiliation{
  	\institution{University of Electronic Science and Technology of China, Megvii Technology}
  	\country{China}}
  \email{liushuaicheng@uestc.edu.cn}

\begin{abstract}
Diffusion models have achieved promising results in image restoration tasks, yet suffer from time-consuming, excessive computational resource consumption, and unstable restoration. To address these issues, we propose a robust and efficient \textbf{D}iffusion-based \textbf{L}ow-\textbf{L}ight image enhancement approach, dubbed DiffLL. Specifically, we present a wavelet-based conditional diffusion model (WCDM) that leverages the generative power of diffusion models to produce results with satisfactory perceptual fidelity. Additionally, it also takes advantage of the strengths of wavelet transformation to greatly accelerate inference and reduce computational resource usage without sacrificing information. To avoid chaotic content and diversity, we perform both forward diffusion and denoising in the training phase of WCDM, enabling the model to achieve stable denoising and reduce randomness during inference. Moreover, we further design a high-frequency restoration module (HFRM) that utilizes the vertical and horizontal details of the image to complement the diagonal information for better fine-grained restoration. Extensive experiments on publicly available real-world benchmarks demonstrate that our method outperforms the existing state-of-the-art methods both quantitatively and visually, and it achieves remarkable improvements in efficiency compared to previous diffusion-based methods. In addition, we empirically show that the application for low-light face detection also reveals the latent practical values of our method. Code is available at https://github.com/JianghaiSCU/Diffusion-Low-Light.
\end{abstract}

%
% The code below should be generated by the tool at
% http://dl.acm.org/ccs.cfm
% Please copy and paste the code instead of the example below.
%


\setcopyright{acmlicensed}
\acmJournal{TOG}
\acmYear{2023} 
\acmVolume{42} 
\acmNumber{6} 
\acmArticle{} 
\acmMonth{12} 
\acmPrice{15.00}
\acmDOI{10.1145/3618373}

% End generated code

\begin{teaserfigure}
    \centering
    \includegraphics[width=\linewidth]{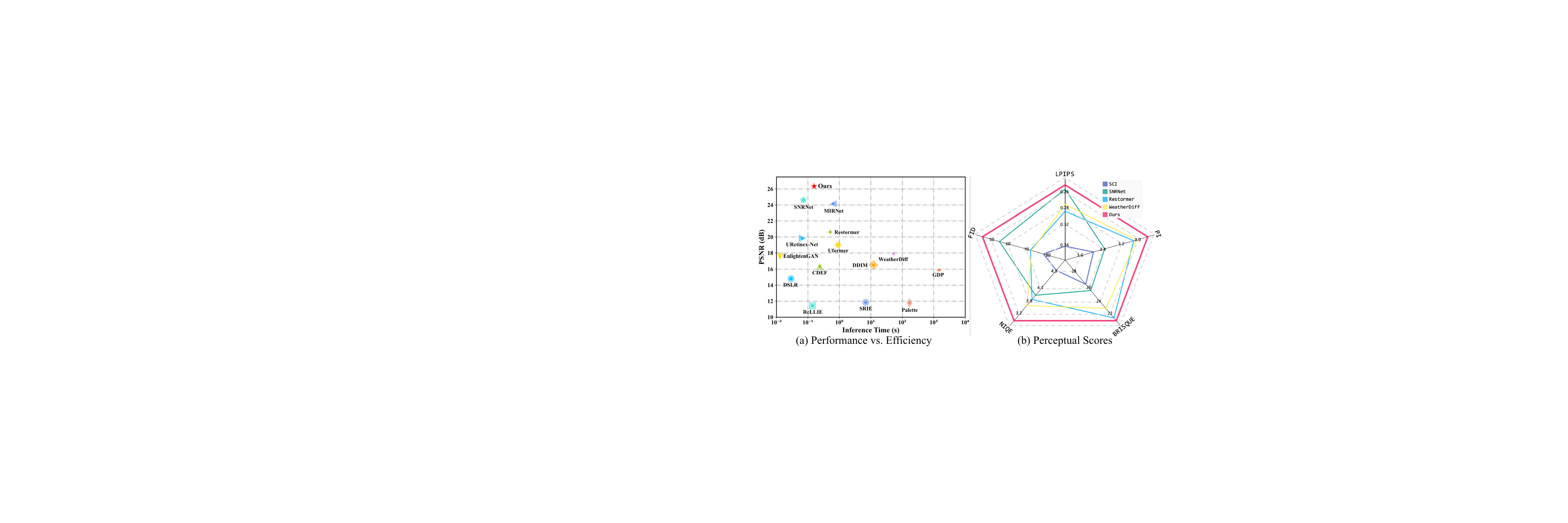}
    \caption{Visual comparisons of recent state-of-the-art unsupervised and supervised low-light image enhancement methods SCI~\cite{SCI} and SNRNet~\cite{SNRNet}, two diffusion-based image restoration methods Palette~\cite{palette} and WeatherDiff~\cite{weatherdiff}, and our method on an image with 2K resolution selected from the UHD-LL~\cite{UHD_ICLR} test set. Previous methods appear incorrect exposure, color distortion, or artifacts to degrade visual quality, while our method properly improves global contrast and presents a vivid color without introducing chaotic content.}
    \label{fig:teaser}
\end{teaserfigure}
\maketitle
\begin{sloppypar}

\section{Introduction}\label{sec:introduction}
Images captured in weakly illuminated conditions suffer from unfavorable visual quality, which seriously impairs the performance of downstream vision tasks and practical intelligent systems, such as image classification~\cite{Classification}, object detection~\cite{Facedetection}, automatic driving~\cite{autonomousdriving}, and visual navigation~\cite{VisualNavigation}, etc. Numerous advances have been made in improving contrast and restoring details to transform low-light images into high-quality ones. Traditional approaches typically rely on optimization-based rules, with the effectiveness of these methods being highly dependent on the accuracy of hand-crafted priors. However, low-light image enhancement (LLIE) is inherently an ill-posed problem, as it presents difficulties in adapting such priors for various illumination conditions. 

With the rise of deep learning, such problems have been partially solved, learning-based methods directly learn mappings between degraded images and corresponding high-quality sharp versions, thus being more robust than traditional methods. Learning-based methods can be divided into two categories: supervised and unsupervised. The former leverage large-scale datasets and powerful neural network architectures for restoration that aim to optimize distortion metrics such as PSNR and SSIM~\cite{SSIM}, while lacking visual fidelity for human perception. Unsupervised methods have better generalization ability for unseen scenes, benefiting from their label-free characteristics. Nevertheless, they are typically unable to control the degree of enhancement and may produce visually unappealing results in some cases, such as over-enhanced or noise amplification, as with supervised ones. As shown in Fig.~\ref{fig:teaser}, both unsupervised SCI~\cite{SCI} and supervised SNRNet~\cite{SNRNet} appear incorrectly overexposed in the background region.

\begin{figure}[!t]
	\centering
	\includegraphics[width=\linewidth]{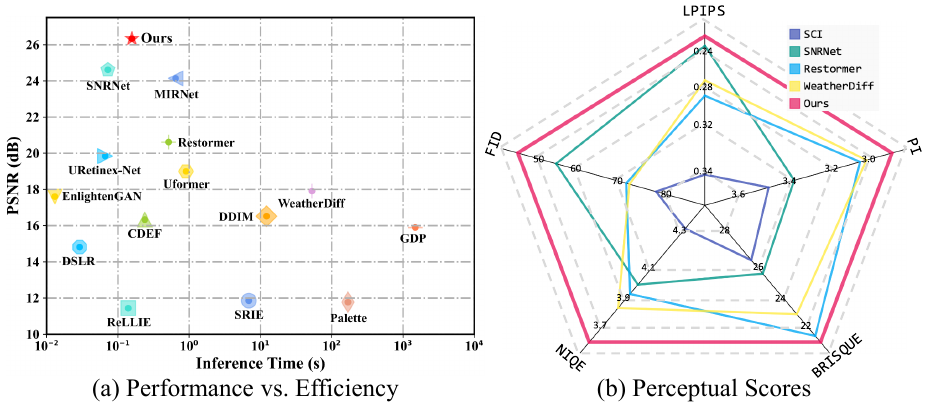}
	\caption{Quantitative comparisons with state-of-the-art methods. (a) presents the performance, i.e., PSNR, and inference time of comparison methods on the LOLv1~\cite{RetinexNet} test set, where the image size is 600$\times$400. (b) presents numerical scores for five perceptual metrics including LPIPS~\cite{LPIPS}, FID~\cite{fid}, NIQE~\cite{NIQE}, BRISQUE~\cite{BRISQUE}, and PI~\cite{PI} (the lower the scores the better the visual quality) on the LOLv1 test set and five real-world benchmarks including DICM~\cite{DICM}, MEF~\cite{MEF}, LIME~\cite{LIME}, NPE~\cite{NPE}, and VV. It can be easily observed that our method is remarkably superior to others, achieving state-of-the-art performance while being computational efficiency.}
	\label{fig:efficiency}
\end{figure}

Therefore, following perceptual-driven approaches~\cite{VGGLoss,perceptual_driven}, deep generative models such as generative adversarial networks (GANs)~\cite{EnlightenGAN,low_light_GAN} and variational autoencoders (VAEs)~\cite{LUD_VAE} have emerged as promising approaches for LLIE. These methods aim to capture multi-modal distributions of inputs to generate output images with better perceptual fidelity. Recently, diffusion models (DMs)~\cite{diffusion_1,diffusion_2,diffusion_3} have garnered attention due to their impressive performance in image synthesis~\cite{ddpm,ddim} and restoration tasks~\cite{Refusion,D2C-SR}. DMs rely on a hierarchical architecture of denoising autoencoders that enable them to iteratively reverse a diffusion process and achieve high-quality mapping from randomly sampled Gaussian noise to target images or latent distributions~\cite{latent_diffusion}, without suffering instability and mode-collapse present in previous generative models. Although standard DMs are capable enough, there exist several challenges for some image restoration tasks, especially for LLIE. As shown in Fig.~\ref{fig:teaser}, the diffusion-based methods Palette~\cite{palette} and WeatherDiff~\cite{weatherdiff} generate images with color distortion or artifacts, since the reverse process starts from a random sampled Gaussian noise, resulting in the final result may have unexpected chaotic content due to the diversity of the sampling process, despite conditional inputs can be used to constrain the output distribution. Moreover, DMs usually require extensive computational resources and long inference times to achieve effective iterative denoising, as shown in Fig.~\ref{fig:efficiency} (a), the diffusion-based methods take more than 10 seconds to restore an image with the size of 600$\times$400.

In this work, we propose a diffusion-based framework, named DiffLL, for robust and efficient low-light image enhancement. Specifically, we first convert the low-light image into the wavelet domain using $K$ times 2D discrete wavelet transformations (2D-DWT), which noticeably reduces the spatial dimension while avoiding information loss, resulting in an average coefficient that represents the global information of the input image and $K$ sets of high-frequency coefficients that represent sparse vertical, horizontal, and diagonal details of the image. Accordingly, we propose a wavelet-based conditional diffusion model (WCDM) that performs diffusion operations on the average coefficient instead of the original image space or latent space for the reduction of computational resource consumption and inference time. As depicted in Fig.~\ref{fig:efficiency} (a), our method achieves a speed that is over \textbf{70$\times$} faster than the previous diffusion-based method DDIM~\cite{ddim}. In addition, during the training phase of the WCDM, we perform both the forward diffusion and the denoising processes. This enables us to fully leverage the generative capacity of the diffusion mode while teaching the model to conduct stable sampling during inference, preventing content diversity even with randomly sampled noise. As shown in Fig.~\ref{fig:teaser}, our method properly improves global contrast and prevents excessive correction on the well-exposed region, without any artifacts or chaotic content appearance. Besides, perceptual scores shown in Fig.~\ref{fig:efficiency} (b) also indicate that our method generates images with better visual quality. Furthermore, the local details contained in the high-frequency coefficients are reconstructed through our well-designed high-frequency restoration modules (HFRM), where the vertical and horizontal information is utilized to complement the diagonal details for better fine-grained details restoration which contributes to the overall quality. Extensive experiments on publicly available datasets show that our method outperforms the existing state-of-the-art (SOTA) methods in terms of both distortion and perceptual metrics, and offers notable improvements in efficiency compared to previous diffusion-based methods. The application for low-light face detection also reveals the potential practical values of our method. Our contributions can be summarized as follows:
\begin{itemize}
	\item We propose a wavelet-based conditional diffusion model (WCDM) that leverages the generative ability of diffusion models and the strengths of wavelet transformation for robust and efficient low-light image enhancement. 
	\item We propose a new training strategy that enables WCDM to achieve content consistency during inference by performing both the forward diffusion and the denoising processes in the training phase. 
	\item We further design a high-frequency restoration module (HFRM) that utilizes both vertical and horizontal information to complement the diagonal details for local details reconstruction.
	\item Extensive experimental results on public real-world benchmarks show that our proposed method achieves state-of-the-art performance on both distortion metrics and perceptual quality while offering noticeable speed up compared to previous diffusion-based methods.
\end{itemize}

\section{Related Work}\label{sec:related_work}
\subsection{Low-light Image Enhancement}
To transform low-light images into visually satisfactory ones, numerous efforts have made considerable advances in improving contrast and restoring details. Traditional methods mainly adopt histogram equalization (HE)~\cite{HE} and Retinex theory~\cite{Retinex} to achieve image enhancement. HE-based methods~\cite{HE1,HE2,HE3} take effect by changing the histogram of the image to improve the contrast. Retinex-based methods~\cite{Retinex_based1,SRIE,Retinex_based2} decompose the image into an illumination map and a reflectance map, by changing the dynamic range of pixels in the illumination map and suppressing the noise in the reflectance map to improve the visual quality of the image. For example, LIME~\cite{LIME} utilized a weighted vibration model based on a prior hypothesis to adjust the estimated illumination, followed by BM3D~\cite{BM3D} as a post-processing step for noise removal.
\begin{figure*}[!t]
	\centering
	\includegraphics[width=\linewidth]{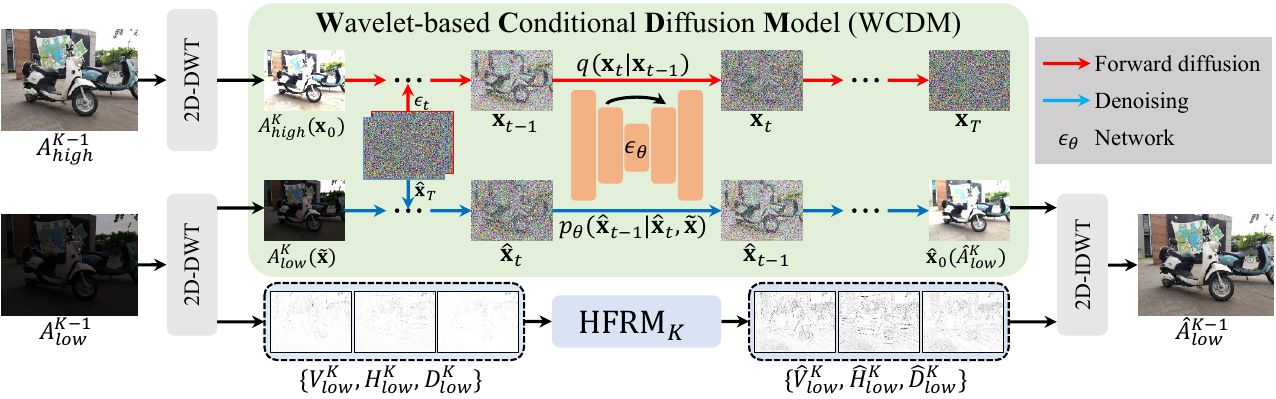}
	\caption{The overall pipeline of our proposed DiffLL. We first convert the low-light image into the wavelet domain using 2D discrete wavelet transformation (2D-DWT) in $K$ times, resulting in an average coefficient $A_{low}^{K}$ and $K$ sets of high-frequency coefficients $\{V_{low}^{k}, H_{low}^{k}, D_{low}^{k}\}$, where $k \in [1, K]$. The proposed wavelet-based conditional model (WCDM) performs diffusion operations on the average coefficient to achieve robust and efficient restoration, in which the training and inference strategies are detailed in ALGORITHM~\ref{algo:1} and ALGORITHM~\ref{algo:2}, respectively. The forward diffusion process is only performed in the training phase and the denoising process is performed in both training and inference phases. Finally, the result at scale $k-1$ is obtained by converting the restored average coefficient and the high-frequency coefficients at scale $k$ using the 2D inverse discrete wavelet transformation (2D-IDWT).}
	\label{fig:pipeline}
\end{figure*}

Following the development of learning-based image restoration methods~\cite{restoration1,restoration2,restoration3}, LLNet~\cite{LLNet} first proposed a deep encoder-decoder network for contrast enhancement and introduced a dataset synthesis pipeline. RetinexNet~\cite{RetinexNet} introduced the first paired dataset captured in real-world scenes and combined Retinex theory with a CNN network to estimate and adjust the illumination map. EnlightenGAN~\cite{EnlightenGAN} adopted unpaired images for training for the first time by using the generative adverse network as the main framework. Zero-DCE~\cite{Zero-DCE} proposed to transform LLIE into a curve estimation problem and designed a zero-reference learning strategy for training. RUAS~\cite{RUAS} proposed a Retinex-inspired architecture search framework to discover low-light prior architectures from a compact search space. SNRNet~\cite{SNRNet} utilized signal-to-noise-ratio-aware transformers and CNN models with spatial-varying operations for restoration. A more comprehensive survey can be found in~\cite{survey}.

\subsection{Diffusion-based Image Restoration}
Diffusion-based generative models have yielded encouraging results with improvements employed in denoising diffusion probability models~\cite{ddpm}, which become more impactful in image restoration tasks such as super-resolution~\cite{IR_SDE,ResDiff}, inpainting~\cite{ddpm_inpainting}, and deblurring~\cite{ddpm_deblurring}. DDRM~\cite{ddrm} used a pre-trained diffusion model to solve any linear inverse problem, which presented superior results over recent unsupervised methods on several image restoration tasks. ShadowDiffusion~\cite{shadowdiffusion} designed an unrolling diffusion model that achieves robust shadow removal by progressively refining the results with degradation and generative priors. DR2~\cite{DR2} first used a pre-trained diffusion model for coarse degradation removal, followed by an enhancement module for finer blind face restoration. WeatherDiff~\cite{weatherdiff} proposed a patch-based diffusion model to restore images captured in adverse weather conditions, which employed a guided denoising process across the overlapping patches in the inference process.

Although diffusion-based methods can obtain restored images with better visual quality than GAN-based and VAE-based generative approaches~\cite{gan_face_restoration,VAE_inpainting}, they often suffer from high computational resource consumption and inference time due to multiple forward and backward passes through the entire network. In this paper, we propose to combine diffusion models with wavelet transformation to noticeably reduce the spatial dimension of inputs in the diffusion process, achieving more efficient and robust low-light image enhancement.

\section{Method}\label{sec:method}
The overall pipeline of our proposed DiffLL is illustrated in Fig. ~\ref{fig:pipeline}. Our approach leverages the generative ability of diffusion models and the strengths of wavelet transformation to achieve visually satisfactory image restoration and improve the efficiency of diffusion models. The wavelet transformation can halve the spatial dimensions after each transformation without sacrificing information, while other transformation techniques, such as Fast Fourier Transformation (FFT) and Discrete Cosine Transform (DCT), are unable to achieve this level of reduction and may result in information loss. Therefore, we conduct diffusion operations on the wavelet domain instead of on the image space. In this section, we begin by presenting the preliminaries of 2D discrete wavelet transformation (2D-DWT) and conventional diffusion models. Then, we introduce the wavelet-based conditional diffusion model (WCDM), which forms the core of our approach. Finally, we present the high-frequency restoration module (HFRM), a well-designed component that contributes to local details reconstruction and improvement in overall quality.
\subsection{Discrete Wavelet Transformation}\label{subsec:Discrete Wavelet Transformation}
\begin{figure}[!t]
	\centering
	\includegraphics[width=\linewidth]{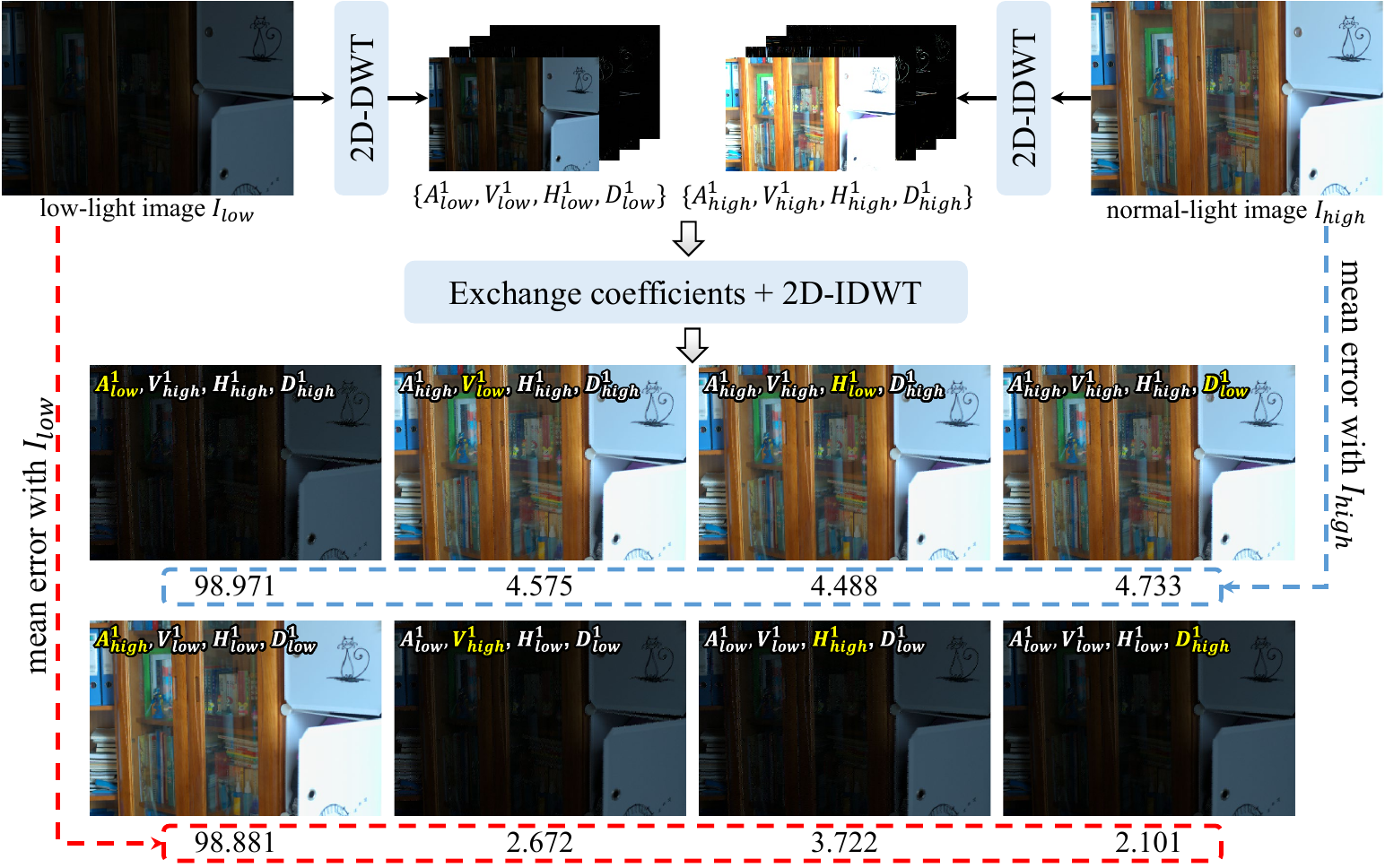}
	\caption{Illustration of exchanging wavelet coefficients between the low-light and normal-light images. Replacing the average coefficient in reconstructed images alters global illumination while exchanging high-frequency coefficients approximately retains the same content as the original images. The quantitative results in mean error also demonstrate that the average coefficient contains richer information than the high-frequency coefficients.}
	\label{fig:Wavelet_combination}
\end{figure}

Given a low-light image $I_{low} \in \mathbb{R}^{H \times W \times c}$, we use 2D discrete wavelet transformation (2D-DWT) with Haar wavelets~\cite{haar} to transform the input into four sub-bands, i.e.,
\begin{equation}\label{eq:1}
	\{A_{low}^{1}, V_{low}^{1}, H_{low}^{1}, D_{low}^{1}\} = \operatorname{2D-DWT}(I_{low}),
\end{equation}
where $A_{low}^{1}, V_{low}^{1}, H_{low}^{1}, D_{low}^{1} \in \mathbb{R}^{\frac{H}{2} \times \frac{W}{2} \times c}$ represent the average of the input image and high-frequency information in the vertical, horizontal, and diagonal directions, respectively. In particular, the average coefficient contains the global information of the original image, which can be treated as the downsampled version of the image, and the other three coefficients contain sparse local details. As shown in Fig.~\ref{fig:Wavelet_combination}, the images reconstructed by exchanging high-frequency coefficients still have approximately the same content as the original images, whereas the image reconstructed by replacing the average coefficient changes the global information, resulting in the largest error with the original images. Therefore, the primary focus on restoring the low-light image in the wavelet domain is to obtain the average coefficient that has natural illumination consistent with its normal-light counterpart. For this purpose, we utilize the generative capability of diffusion models to restore the average coefficient, and the remaining three high-frequency coefficients are reconstructed through the proposed HFRM to facilitate local details restoration.

Although the spatial dimension of the wavelet component processed by the diffusion model after one wavelet transformation is four times smaller than the original image, we further perform $K-1$ times wavelet transformations on the average coefficient, i.e.,
\begin{equation}\label{eq:2}
	\{A_{low}^{k}, V_{low}^{k}, H_{low}^{k}, D_{low}^{k}\} = \operatorname{2D-DWT}(A_{low}^{k-1}),
\end{equation}
where $A_{low}^{k}, V_{low}^{k}, H_{low}^{k}, D_{low}^{k} \in \mathbb{R}^{\frac{H}{2^{k}} \times \frac{W}{2^{k}} \times c}$, $k \in [1, K]$,  and the original input image could be denoted as $A_{low}^{0}$. Subsequently, we do diffusion operations on the $A_{low}^{K}$ to further improve the efficiency, and the high-frequency coefficients $\{V_{low}^{k}, H_{low}^{k}, D_{low}^{k}\}$ are also reconstructed by the HFRM$_{k}$. In this way, our approach achieves significant decreases in inference time and computational resource consumption of the diffusion model due to $4^{K}$ times reduction in the spatial dimension.

\subsection{Conventional Conditional Diffusion Models}\label{subsec:Conventional Conditional Diffusion Models}
Diffusion models~\cite{ddim,ddpm} aim to learn a Markov Chain to gradually transform the distribution of Gaussian noise into the training data, which is generally divided into two phases: the forward diffusion and the denoising. The forward diffusion process first uses a fixed variance schedule $\{\beta_1, \beta_2, \cdots, \beta_T\}$ to progressively transform the input $\mathbf{x}_0$ into corrupted noise data $\mathbf{x}_T \sim \mathcal{N}(\mathbf{0},\mathbf{I})$ through $T$ steps, which can be formulated as:
\begin{equation}\label{eq:3}
	q(\mathbf{x}_{1:T} \mid \mathbf{x}_0)=\prod_{t=1}^T q(\mathbf{x}_t \mid \mathbf{x}_{t-1}),
\end{equation}
\begin{equation}\label{eq:4}
	q(\mathbf{x}_t \mid \mathbf{x}_{t-1})=\mathcal{N}(\mathbf{x}_t ; \sqrt{1-\beta_t} \mathbf{x}_{t-1}, \beta_t \mathbf{I}),
\end{equation}
where $\mathbf{x}_t$ and $\beta_t$ are the corrupted noise data and the pre-defined variance at time-step $t$, respectively, and $\mathcal{N}$ represents the Gaussian distribution.

The inverse process learns the Gaussian denoising transitions starting from the standard normal prior $p(\mathbf{\hat{x}}_T)=\mathcal{N}(\mathbf{\hat{x}}_T;\mathbf{0},\mathbf{I})$, and gradually denoise a randomly sampled Gaussian noise $\mathbf{\hat{x}}_T \sim \mathcal{N}(\mathbf{0},\mathbf{I})$ into a sharp result $\mathbf{\hat{x}}_0$, i.e.,
\begin{equation}\label{eq:5}
	p_\theta(\mathbf{\hat{x}}_{0:T})=p(\mathbf{\hat{x}}_T) \prod_{t=1}^T p_\theta(\mathbf{\hat{x}}_{t-1} \mid \mathbf{\hat{x}}_t).
\end{equation}
By applying the editing and data synthesis capabilities of conditional diffusion models~\cite{conditional_ddpm}, we aim to learn conditional denoising process $p_\theta(\mathbf{\hat{x}}_{0:T} \mid \tilde{\mathbf{x}})$ without changing the forward diffusion for $\mathbf{x}$, which results in high fidelity of the sampled result to the distribution conditioned on the conditional input $\tilde{\mathbf{x}}$. The Eq.\ref{eq:5} is converted to:
\begin{equation}\label{eq:6}
	p_\theta(\mathbf{\hat{x}}_{0:T} \mid \tilde{\mathbf{x}})=p(\mathbf{\hat{x}}_T) \prod_{t=1}^T p_\theta(\mathbf{\hat{x}}_{t-1} \mid \mathbf{\hat{x}}_t, \tilde{\mathbf{x}}),
\end{equation}
\begin{equation}\label{eq:7}
	p_\theta(\mathbf{\hat{x}}_{t-1} \mid \mathbf{\hat{x}}_t)=\mathcal{N}(\mathbf{\hat{x}}_{t-1};\boldsymbol{\mu}_\theta(\mathbf{\hat{x}}_t, t), \sigma_t^2 \mathbf{I}).
\end{equation}
The training objective for diffusion models is to optimize a network $\boldsymbol{\mu}_\theta(\mathbf{x}_t, \tilde{\mathbf{x}}, t)=\frac{1}{\sqrt{\alpha_t}}(\mathbf{x}_t-\frac{\beta_t}{\sqrt{1-\bar{\alpha}_t}} \boldsymbol{\epsilon}_\theta(\mathbf{x}_t, \tilde{\mathbf{x}}, t))$ that predicts $\tilde{\boldsymbol{\mu}}_t$, where the model can predict the noise vector $\boldsymbol{\epsilon}_\theta(\mathbf{x}_t, \tilde{\mathbf{x}}, t)$ by optimizing the network parameters $\theta$ of $\boldsymbol{\epsilon}_\theta$ like~\cite{ddpm}. The objective function is formulated as:
\begin{equation}\label{eq:8}
	\mathcal{L}_{diff} = E_{\mathbf{x}_0, t, \epsilon_t \sim \mathcal{N}(\mathbf{0},\mathbf{I})}[\|\epsilon_t-\epsilon_\theta(\mathbf{x}_t, \tilde{\mathbf{x}}, t)\|^2].
\end{equation}

\subsection{Wavelet-based Conditional Diffusion Models}\label{subsec:Wavelet-based Conditional Diffusion Models}
However, for some image restoration tasks, there are two challenges with the above conventional diffusion and reverse processes: 1) The forward diffusion with large time step $T$, typically set to $T=1000$, and small variance $\beta_t$ permits the assumption that the denoising process becomes close to Gaussian, which results in costly inference time and computational resources consumption. 2) Since the reverse process starts from a randomly sampled Gaussian noise, the diversity of the sampling process may lead to content inconsistency in the restored results, despite the conditioned input enforcing constraints on the distribution of the sampling results.
\begin{algorithm}[!t]
	\caption{Wavelet-based conditional diffusion model training}
	\label{algo:1} 
	\SetKwData{Left}{left}\SetKwData{This}{this}\SetKwData{Up}{up} \SetKwFunction{Union}{Union}\SetKwFunction{FindCompress}{FindCompress} \SetKwInOut{Input}{input}\SetKwInOut{Output}{output}
	
	\Input{Average coefficients of low/normal-light image pairs $A_{low}^{K}$ and $A_{high}^{K}$, denoted as $\tilde{\mathbf{x}}$ and $\mathbf{x}_0$, respectively, the time step $T$, and the number of implicit sampling step $S$.} 
	
	\While{Not converged}{
		
		\emph{\# Forward diffusion process}
		
		$t \sim \operatorname{Uniform}\{1,\cdots,T\}$
		
		$\epsilon_t \sim \mathcal{N}(\mathbf{0},\mathbf{I})$ 
		
		Perform a single gradient descent step for
		$\nabla_\theta\|\boldsymbol{\epsilon}_t-\boldsymbol{\epsilon}_\theta(\sqrt{\bar{\alpha}_t} \mathbf{x}_0+\sqrt{1-\bar{\alpha}_t} \boldsymbol{\epsilon}_t, \tilde{\mathbf{x}}, t)\|^2$
		
		\emph{\# Denoising process}
		
		$\mathbf{\hat{x}}_T \sim \mathcal{N}(\mathbf{0},\mathbf{I})$
		
		\For{$i = S:1$}{ 
			$t = (i-1) \cdot T/S + 1$
			
			$t_{\operatorname{next}} = (i-2) \cdot T/S + 1$ if $i > 1$, else $0$
			
			$\mathbf{\hat{x}}_t \leftarrow \sqrt{\bar{\alpha}_{t_{\operatorname{next}}}}(\frac{\mathbf{\hat{x}}_t-\sqrt{1-\bar{\alpha}_t} \cdot \epsilon_\theta(\mathbf{\hat{x}}_t, \tilde{\mathbf{x}}, t)}{\sqrt{\bar{\alpha}_t}}) +\sqrt{1-\bar{\alpha}_{t_{\operatorname{next}}}} \cdot \epsilon_\theta(\mathbf{\hat{x}}_t, \tilde{\mathbf{x}}, t)$
		}
		
		Perform a single gradient descent step for
		$\nabla_\theta\|\mathbf{\hat{x}}_0 - \mathbf{x}_0\|^2$}
	
	\Output{$\theta$, $\mathbf{\hat{x}}_0$ ($\hat{A}_{low}^{K}$)}
\end{algorithm}
\begin{algorithm}[!t]
	\caption{Wavelet-based conditional diffusion model inference}
	\label{algo:2} 
	\SetKwData{Left}{left}\SetKwData{This}{this}\SetKwData{Up}{up} \SetKwFunction{Union}{Union}\SetKwFunction{FindCompress}{FindCompress} \SetKwInOut{Input}{input}\SetKwInOut{Output}{output}
	
	\Input{Average coefficients of low-light image $A_{low}^{k}$, denoted as $\tilde{\mathbf{x}}$, the time step $T$, and the number of implicit sampling step $S$.} 
	\emph{\# Denoising process}
	
	$\mathbf{\hat{x}}_T \sim \mathcal{N}(\mathbf{0},\mathbf{I})$
	
	\For{$i = S:1$}{ 
		$t = (i-1) \cdot T/S + 1$
		
		$t_{\operatorname{next}} = (i-2) \cdot T/S + 1$ if $i > 1$, else $0$
		
		$\mathbf{\hat{x}}_t \leftarrow \sqrt{\bar{\alpha}_{t_{\operatorname{next}}}}(\frac{\mathbf{\hat{x}}_t-\sqrt{1-\bar{\alpha}_t} \cdot \epsilon_\theta(\mathbf{\hat{x}}_t, \tilde{\mathbf{x}}, t)}{\sqrt{\bar{\alpha}_t}}) +\sqrt{1-\bar{\alpha}_{t_{\operatorname{next}}}} \cdot \epsilon_\theta(\mathbf{\hat{x}}_t, \tilde{\mathbf{x}}, t)$    
	}
	\Output{$\mathbf{\hat{x}}_0$ ($\hat{A}_{low}^{k}$)}
\end{algorithm}

To address these problems, we propose a wavelet-based conditional diffusion model (WCDM) that converts the input low-light image $I_{low}$ into the wavelet domain and performs the diffusion process on the average coefficient, which considerably reduces the spatial dimension for efficient restoration. In addition, we perform both the forward diffusion and the denoising processes in the training phase, which aids the model in achieving stable sampling during inference and avoiding content diversity. The training approach of our wavelet-based conditional diffusion model is summarized in Algorithm~\ref{algo:1}. Specifically, we first perform the forward diffusion to optimize a noise estimator network via Eq.\ref{eq:8}, then adopt the sampled noise $\mathbf{\hat{x}}_T$ with conditioned input $\tilde{\mathbf{x}}$, i.e., the average coefficient $A_{low}^{K}$, for the denoising process, resulting in the restored coefficient $\hat{A}_{low}^{K}$. The content consistency is realized by minimizing the L2 distance between the restored coefficient and the reference coefficient $A_{high}^{K}$ which is only available during the training phase. Accordingly, the objective function Eq.\ref{eq:8} used to optimize the diffusion model is rewritten as:
\begin{equation}\label{eq:9}
	\mathcal{L}_{diff} = E_{\mathbf{x}_0, t, \epsilon_t \sim \mathcal{N}(\mathbf{0},\mathbf{I})}[\|\epsilon_t-\epsilon_\theta(\mathbf{x}_t, \tilde{\mathbf{x}}, t)\|^2] + \|\hat{A}_{low}^{K} - A_{high}^{K}\|^2.
\end{equation}
\begin{figure}[!t]
	\centering
	\includegraphics[width=\linewidth]{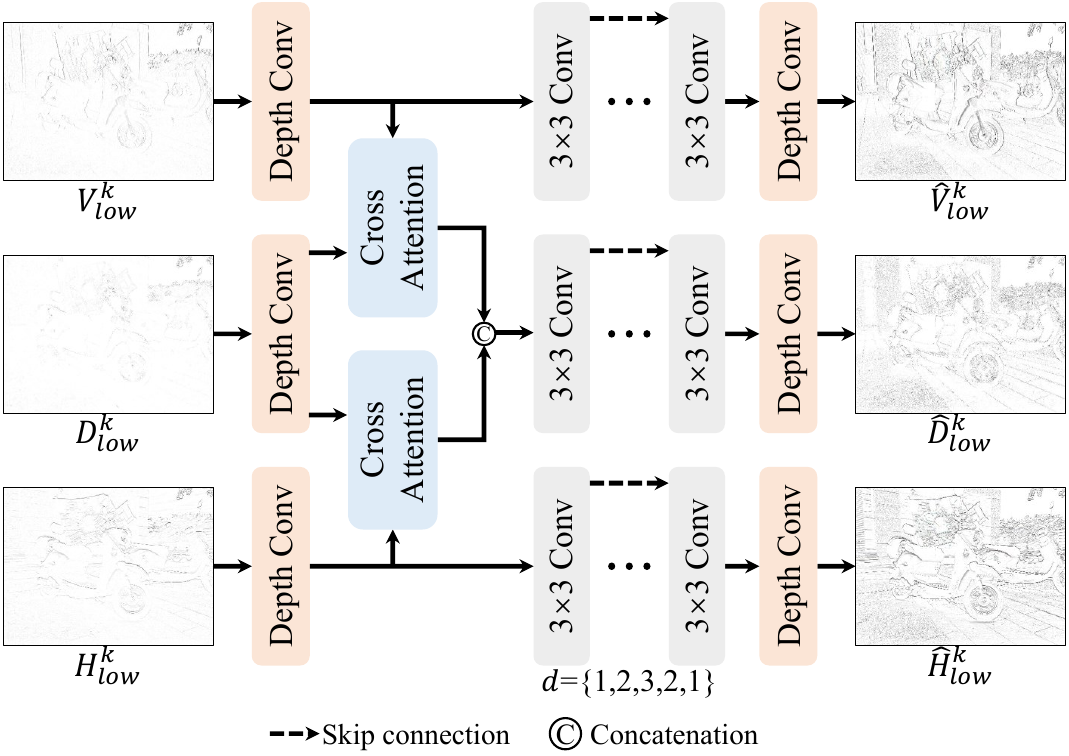}
	\caption{The detailed architecture of our proposed high-frequency restoration module. Depth Conv denotes the depth-wise separable convolution.}
	\label{fig:HFRM_architecture}
\end{figure}

{\textbf{Discussion.} There exist some diffusion models, such as Latent Diffusion~\cite{latent_diffusion}, that adopt VAE to transform the original image into the feature level and perform diffusion operations in the latent space, which can also achieve spatial dimension reduction as our method. However, encoding the original image through the VAE encoder inevitably causes information loss, whereas the wavelet transformation offers the advantage of preserving all information without any sacrifice. In addition, utilizing VAE would introduce an increase in model parameters, whereas the wavelet transformation, being a linear operation, does not incur additional resource consumption. On the other hand, the data typically used for training VAE are captured under normal-light conditions, it should be retrained on the low-light dataset to prevent domain shifts and maintain effective generation capability.}

\subsection{High-Frequency Restoration Module}\label{subsec:High-Frequency Restoration Module}
For the $k$-th level wavelet sub-bands of the low-light image, the high-frequency coefficients $V_{low}^{k}$, $H_{low}^{k}$, and $D_{low}^{k}$ hold sparse representations of vertical, horizontal, and diagonal details of the image. To restore the low-light image to contain as rich information as the normal-light image, we propose a high-frequency restoration module (HFRM) to reconstruct these coefficients. As shown in Fig.~\ref{fig:HFRM_architecture}, we first use three depth-wise separable convolutions~\cite{depth_conv} for the sake of efficiency to extract the features of input coefficients, then two cross-attention layers~\cite{cross_attention} are employed to leverage the information in $V$ and $H$ to complement the details in $D$. Subsequently, inspired by~\cite{SFDNet}, we design a progressive dilation Resblock utilizing dilation convolutions for local restoration, where the first and last convolutions are utilized to extract local information, and the middle dilation convolutions are used to improve the receptive field to make better use of long-range information. By gradually increasing and decreasing the dilation rate $d$, the gridding effect can be avoided. Finally, three depth-wise separable convolutions are used to reduce channels to obtain the reconstructed $\hat{V}_{low}^{k}$, $\hat{H}_{low}^{k}$, and $\hat{D}_{low}^{k}$. The restored average coefficient and the high-frequency coefficients at scale $k$ are correspondingly converted into the output at scale $k-1$ by employing the 2D inverse discrete wavelet transformation (2D-IDWT), i.e.,
\begin{equation}\label{eq:10}
	\hat{A}_{low}^{k-1} = \operatorname{2D-IDWT}(\{\hat{A}_{low}^{k}, \hat{V}_{low}^{k}, \hat{H}_{low}^{k}, \hat{D}_{low}^{k}\}),
\end{equation}
where the $\hat{A}_{low}^{0}$ denotes the final restored image $\hat{I}_{low}$.
\begin{table*}[!t]
	\centering
	\caption{Quantitative evaluation of different methods on LOL-v1~\cite{RetinexNet}, LOLv2-real~\cite{LOLV2}, and LSRW~\cite{R2RNet} test sets. The best results are highlighted in \textbf{bold} and the second best results are \underline{underlined}. $\ast$ denotes the methods are retrained on the LOLv1 training set.}
	\resizebox{\linewidth}{!}{
		\begin{tabular}{l|c|cccc|cccc|cccc}
			\toprule
			\multirow{2}[4]{*}{Methods} & \multirow{2}[4]{*}{Reference} & \multicolumn{4}{c|}{LOLv1} & \multicolumn{4}{c|}{LOLv2-real} & \multicolumn{4}{c}{LSRW} \\
			\cmidrule{3-14} & & PSNR $\uparrow$ & SSIM $\uparrow$ & LPIPS $\downarrow$ & FID $\downarrow$ & PSNR $\uparrow$ & SSIM $\uparrow$ & LPIPS $\downarrow$ & FID $\downarrow$ & PSNR $\uparrow$ & SSIM $\uparrow$ & LPIPS $\downarrow$ & FID $\downarrow$ \\
			\midrule
			NPE & TIP' 13 & 16.970 & 0.484 & 0.400 & 104.057 & 17.333 & 0.464 & 0.396 & 100.025 & 16.188 & 0.384 & 0.440 & 90.132 \\
			SRIE & CVPR' 16 & 11.855 & 0.495 & 0.353 & 88.728 & 14.451 & 0.524 & 0.332 & 78.834 & 13.357 & 0.415 & \underline{0.399} & 69.082 \\
			LIME & TIP' 16 & 17.546 & 0.531 & 0.387 & 117.892 & 17.483 & 0.505 & 0.428 & 118.171 & 17.342 & 0.520 & 0.471 & 75.595 \\
			RetinexNet & BMVC' 18 & 16.774 & 0.462 & 0.417 & 126.266 & 17.715 & 0.652 & 0.436 & 133.905 & 15.609 & 0.414 & 0.454 & 108.350 \\
			DSLR & TMM' 20 & 14.816 & 0.572 & 0.375 & 104.428 & 17.000 & 0.596 & 0.408 & 114.306 & 15.259 & 0.441 & 0.464 & 84.930 \\
			DRBN & CVPR' 20 & 16.677 & 0.730 & 0.345 & 98.732 & 18.466 & 0.768 & 0.352 & 89.085 & 16.734 & 0.507 & 0.457 & 80.727 \\
			Zero-DCE & CVPR' 20 & 14.861 & 0.562 & 0.372 & 87.238 & 18.059 & 0.580 & 0.352 & 80.449 & 15.867 & 0.443 & 0.411 & 63.320 \\
			MIRNet & ECCV' 20 & 24.138 & 0.830 & 0.250 & 69.179 & 20.020 & 0.820 & 0.233 & \underline{49.108} & 16.470 & 0.477 & 0.430 & 93.811 \\
			EnlightenGAN & TIP' 21 & 17.606 & 0.653 & 0.372 & 94.704 & 18.676 & 0.678 & 0.364 & 84.044 & 17.106 & 0.463 & 0.406 & 69.033 \\
			ReLLIE & ACM MM' 21& 11.437 & 0.482 & 0.375 & 95.510 & 14.400 & 0.536 & 0.334 & 79.838 & 13.685 & 0.422 & 0.404 & 65.221 \\
			RUAS & CVPR' 21 & 16.405 & 0.503 & 0.364 & 101.971 & 15.351 & 0.495 & 0.395 & 94.162 & 14.271 & 0.461 & 0.501 & 78.392 \\
			DDIM$^{\ast}$ & ICLR' 21 & 16.521 & 0.776 & 0.376 & 84.071 & 15.280 & 0.788 & 0.387 & 76.387 & 14.858 & 0.486 & 0.495 & 71.812 \\
			CDEF & TMM' 22 & 16.335 & 0.585 & 0.407 & 90.620 & 19.757 & 0.630 & 0.349 & 74.055 & 16.758 & 0.465 & \underline{0.399} & 62.780 \\
			SCI & CVPR' 22 & 14.784 & 0.525 & 0.366 & 78.598 & 17.304 & 0.540 & 0.345 & 67.624 & 15.242 & 0.419 & 0.404 & \underline{56.261} \\
			URetinex-Net & CVPR' 22 & 19.842 & 0.824 & 0.237 & \underline{52.383} & 21.093 & \underline{0.858} & \underline{0.208} & 49.836 & \underline{18.271} & 0.518 & 0.419 & 66.871 \\
			SNRNet & CVPR' 22 & \underline{24.610} & \underline{0.842} & \underline{0.233} & 55.121 & 21.480 & 0.849 & 0.237 & 54.532 & 16.499 & 0.505 & 0.419 & 65.807 \\
			Uformer$^{\ast}$ & CVPR' 22 & 19.001 & 0.741 & 0.354 & 109.351 & 18.442 & 0.759 & 0.347 & 98.138 & 16.591 & 0.494 & 0.435 & 82.299 \\
			Restormer$^{\ast}$ & CVPR' 22 & 20.614 & 0.797 & 0.288 & 72.998 & \underline{24.910} & 0.851 & 0.264 & 58.649 & 16.303 & 0.453 & 0.427 & 69.219 \\
			Palette$^{\ast}$ & SIGGRAPH' 22 & 11.771 & 0.561 & 0.498 & 108.291 & 14.703 & 0.692 & 0.333 & 83.942 & 13.570 & 0.476 & 0.479 & 73.841 \\
			UHDFour$_{2\times}$ & ICLR' 23 & 23.093 & 0.821 & 0.259 & 56.912 & 21.785 & 0.854 & 0.292 & 60.837 & 17.300 & \underline{0.529} & 0.443 & 62.032 \\
			%\cdashline{1-14}[2pt/1pt]
			WeatherDiff$^{\ast}$ & TPAMI' 23 & 17.913 & 0.811 & 0.272 & 73.903 & 20.009 & 0.829 & 0.253 & 59.670 & 16.507 & 0.487 & 0.431 & 96.050 \\
			GDP & CVPR' 23 & 15.896 & 0.542 & 0.421 & 117.456 & 14.290 & 0.493 & 0.435 & 102.416 & 12.887 & 0.362 & 0.412 & 76.908 \\
			DiffLL(Ours) & - & \textbf{26.336} & \textbf{0.845} & \textbf{0.217} & \textbf{48.114} & \textbf{28.857} & \textbf{0.876} & \textbf{0.207} & \textbf{45.359} & \textbf{19.281} & \textbf{0.552} & \textbf{0.350} & \textbf{45.294} \\
			\bottomrule
	\end{tabular}}
	\label{tab:evaluation_paired}
\end{table*}

\subsection{Network Training}\label{subsec:Network Training}
Besides the objective function $\mathcal{L}_{diff}$ used to optimize the diffusion model, we also use a detail preserve loss $\mathcal{L}_{detail}$ combing MSE loss and TV loss~\cite{TV} similar to~\cite{RUAS} to reconstruct the high-frequency coefficients, i.e.,
\begin{equation}\label{eq:11}
	\begin{aligned}
		\mathcal{L}_{detail} &= \lambda_{1} \sum_{k=1}^{K} \|\{\hat{V}_{low}^{k}, \hat{H}_{low}^{k}, \hat{H}_{low}^{k}\} - \{V_{high}^{k}, H_{high}^{k}, D_{high}^{k}\}\|^2 \\& + \lambda_{2} \sum_{k=1}^{K} \operatorname{TV}(\{\hat{V}_{low}^{k}, \hat{H}_{low}^{k}, \hat{H}_{low}^{k}\}),
	\end{aligned}
\end{equation}
where $\lambda_{1}$ and $\lambda_{2}$ are the weights of each term set as 0.1 and 0.01, respectively. Moreover, we utilize a content loss $\mathcal{L}_{content}$ that combines L1 loss and SSIM loss~\cite{SSIM} to minimize the content difference between the restored image $\hat{I}_{low}$ and the reference image $I_{high}$, i.e.,
\begin{equation}\label{eq:12}
	\mathcal{L}_{content} = |\hat{I}_{low} - I_{high}|_{1} + (1 - \operatorname{SSIM}(\hat{I}_{low}, I_{high})).
\end{equation}
The total loss $\mathcal{L}_{total}$ is expressed by combing the diffusion objective function, the detail preserve loss, and the content loss as:
\begin{equation}\label{eq:13}
	\mathcal{L}_{total} = \mathcal{L}_{diff} + \mathcal{L}_{detail} + \mathcal{L}_{content}.
\end{equation}

\section{Experiment}\label{sec:experiment}
\subsection{Experimental Settings}\label{subsec:Experimental Settings}
\textbf{Implementation Details.} Our implementation is done with PyTorch. The proposed network can be converged after being trained for $1\times10^{5}$ iterations on four NVIDIA RTX 2080Ti GPUs. The Adam optimizer~\cite{Adam} is adopted for optimization. The initial learning rate is set to $1\times10^{-4}$ and decays by a factor of 0.8 after every $5\times10^{3}$ iterations. The batch size and patch size are set to 12 and 256$\times$256, respectively. The wavelet transformation scale $K$ is set to 2. For our wavelet-based conditional diffusion model, the commonly used U-Net architecture~\cite{Unet} is adopted as the noise estimator network. To achieve efficient restoration, the time step $T$ is set to 200 for the training phase, and the implicit sampling step $S$ is set to 10 for both the training and inference phases. 

\textbf{Datasets.} Our network is trained and evaluated on the LOLv1 dataset~\cite{RetinexNet}, which contains 1,000 synthetic low/normal-light image pairs and 500 real-word pairs, where 15 real-world pairs are adopted for evaluation and the remaining pairs are used for training. Additionally, we adopt 2 real-world paired datasets, including LOLv2-real~\cite{LOLV2} and LSRW~\cite{R2RNet}, to evaluate the performance of the proposed method. More specifically, the LSRW dataset contains 5,650 image pairs captured in various scenes, 5,600 image pairs are randomly selected as the training set and the remaining 50 image pairs are used for evaluation. The LOLv2-real dataset contains 789 image pairs, of which 689 pairs are for training and 100 pairs for evaluation. To validate the effectiveness of our method for high-resolution image restoration, we adopt the UHD-LL dataset~\cite{UHD_ICLR} for evaluation which contains 2,150 low/normal-light Ultra-High-Definition (UHD) image pairs with 4K resolution, where 150 pairs are used for evaluation and the rest for training. Furthermore, we evaluate the generalization ability of the proposed method to unseen scenes on five commonly used real-world unpaired benchmarks, including DICM~\cite{DICM}, LIME~\cite{LIME}, NPE~\cite{NPE}, MEF~\cite{MEF}, and VV\footnote{https://sites.google.com/site/vonikakis/datasets}.

\textbf{Metrics.} For the paired datasets, we adopt two full-reference distortion metrics PSNR and SSIM~\cite{SSIM} to evaluate the performance of the proposed method, and also two perceptual metrics LPIPS~\cite{LPIPS} and FID~\cite{fid} to measure the visual quality of the enhanced results. For the other five unpaired datasets, we use three non-reference perceptual metrics NIQE~\cite{NIQE}, BRISQUE~\cite{BRISQUE}, and PI~\cite{PI} for evaluation.

\subsection{Comparison with Existing Methods}\label{subsec:Comparison with Existing Methods}
\textbf{Comparison Methods.} In this section, we compare our proposed DiffLL with four categories of existing state-of-the-art methods, including 1) optimization-based LLIE methods NPE~\cite{NPE}, SRIE~\cite{SRIE}, LIME~\cite{LIME}, and CDEF~\cite{CDEF}, 2) learning-based LLIE methods RetinexNet~\cite{RetinexNet}, DSLR~\cite{DSLR}, DRBN~\cite{DRBN}, Zero-DCE~\cite{Zero-DCE}, MIRNet~\cite{MIRNet}, EnlightenGAN~\cite{EnlightenGAN}, ReLLIE~\cite{ReLLIE}, RUAS~\cite{RUAS},  SCI~\cite{SCI}, URetinex-Net~\cite{Uretinex-net}, SNRNet~\cite{SNRNet}, and UHDFour~\cite{UHD_ICLR}, 3) transformer-based image restoration methods Uformer~\cite{Uformer} and Restormer~\cite{Restormer}, and 4) diffusion-based methods conditional DDIM~\cite{ddim}, Palette~\cite{palette}, WeatherDiff~\cite{weatherdiff}, and GDP~\cite{GDP}. For fair comparisons, we retrain the transformer-based and diffusion-based methods, except GDP, on the LOLv1 training set using their default implementations, denoted by $^{\ast}$. UHDFour provides two versions for low-resolution and UHD LLIE, which downsample the original input image by a factor of 2 and 8 for efficient restoration, denoted as UHDFour$_{2\times}$ and UHDFour$_{8\times}$, and are trained on the LOLv1 and UHD-LL training sets, respectively. Therefore, we mainly report the performance of the UHDFour$_{2\times}$ version for fair comparisons. Furthermore, the metrics of the diffusion-based methods and our method are the mean values for five times evaluation.

\textbf{Quantitative Comparison.} We first compare our method with all comparison methods on the LOLv1~\cite{RetinexNet}, LOLv2-real~\cite{LOLV2}, and LSRW~\cite{R2RNet} test sets. As shown in Table~\ref{tab:evaluation_paired}, our method achieves SOTA quantitative performance compared with all comparison methods. Specifically, for the distortion metrics, our method achieves 1.726dB(=26.336-24.610) and 0.003(=0.845-0.842) improvements in terms of PSNR and SSIM on the LOLv1 test set compared with the second-best method SNRNet. On the LOLv2-real test set, our DiffLL obtains PSNR and SSIM of 28.857dB and 0.876 that greatly outperform the second-best approaches, i.e., Restormer and URetinex-Net, by 3.947dB(=28.857-24.910) and 0.018(0.876-0.858). On the LSRW test set, our method improves the two metrics by at least 1.01dB(=19.281-18.271) and 0.023(0.552-0.529), respectively. For perceptual metrics, i.e., LPIPS and FID, where regression-based methods and previous diffusion-based methods perform unfavorably, our method yields the lowest perceptual scores on all three datasets, which demonstrates our proposed wavelet-based diffusion model can produce restored images with satisfactory visual quality and is more appropriate for the LLIE task. In particular, for FID, most methods are unable to obtain stable metrics due to the cross-domain issue between training and test data, while on the contrary, our method obtains FID scores less than 50 on all three datasets, proving that our method can generalize well to unknown datasets.
\begin{table}[!t]
	\centering
	\caption{The average time (seconds) and GPU memory (G) costs of different methods consumed on the image with the size of 600$\times$400, 1920$\times$1080 (1080P), and 2560$\times$1440 (2K), respectively, during inference. OOM denotes the out-of-memory error and `-’ denotes unavailable.}
	\resizebox{\linewidth}{!}{
		\begin{tabular}{l|cc|cc|cc}
			\toprule
			\multirow{2}[2]{*}{Methods} & \multicolumn{2}{c|}{600$\times$400} & \multicolumn{2}{c|}{1920$\times$1080 (1080P)} & \multicolumn{2}{c}{2560$\times$1440 (2K)} \\
			& Mem.(G) $\downarrow$ & Time (s) $\downarrow$ & Mem.(G) $\downarrow$ & Time (s) $\downarrow$ & Mem.(G) $\downarrow$ & Time (s) $\downarrow$ \\
			\midrule
			MIRNet & 2.699 & 0.643 & OOM & - & OOM & - \\
			URetinex-Net & 1.656 & 0.066 & 5.416 & 0.454 & 8.699 & 0.814 \\
			SNRNet & 1.523 & 0.072 & 8.125 & 0.601 & OOM & - \\
			Uformer & 4.457 & 0.901 & OOM & - & OOM & - \\
			Restormer & 4.023 & 0.513 & OOM & - & OOM & - \\
			DDIM & 5.881 & 12.138 & OOM & - & OOM & - \\
			Palette & 6.830 & 168.515 & OOM & - & OOM & - \\
			UHDFour$_{2\times}$ & 2.228 & 0.062 & 8.263 & 0.533 & OOM & - \\
			UHDFour$_{8\times}$ & 1.423 & 0.011 & 2.017 & 0.057 & 4.956 & 0.105 \\
			WeatherDiff & 6.570 & 52.703 & 8.344 & 548.708 & OOM & - \\
			DiffLL(Ours) & 1.850 & 0.157 & 3.873 & 1.072 & 7.141 & 2.403 \\
			\bottomrule
	\end{tabular}}
	\label{tab:inference_time}
\end{table}

Besides the enhanced visual quality, efficiency is also an important indicator for image restoration tasks. We report the average time and GPU memory costs of competitive methods and our method during inference on the LOLv1 test set in Table~\ref{tab:inference_time}, where the image size is 600$\times$400. All methods are tested on an NVIDIA RTX 2080Ti GPU. We can see that our method achieves a better trade-off in efficiency and performance, averagely spending 0.157s and 1.850G memory on each image which is at a moderate level in all compared methods. Benefiting from the proposed wavelet-based diffusion model, our method is at least 70$\times$ faster than the previous diffusion-based restoration approaches and consumes at least 3$\times$ fewer computational resources. We also report the evaluation of the inference time and GPU memory consumed on the image with the size of 1920$\times$1080 (1080P) and 2560$\times$1440 (2K), where most comparison methods encounter out-of-memory errors, especially at 2K resolution. In contrast, our method spends much fewer GPU resources, which proves the potential value of our approach for large-resolution LLIE tasks.
\begin{table}[!t]
	\centering
	\caption{Quantitative evaluation of different methods on the UHD-LL~\cite{UHD_ICLR} test set. The best results are highlighted in \textbf{bold} and the second best results are \underline{underlined}. Note that UHDFour$_{2\times}$ is trained on the LOLv1~\cite{RetinexNet} training set like our method, and the methods indicated with ${\ast}$ are also retrained on it. On the other hand, UHDFour$_{8\times}$ is trained on the UHD-LL training set, which brings it stronger capability to handle ultra-high-definition image restoration. The resolution indicates the maximum size that the models can handle during inference.}
	\resizebox{0.88\linewidth}{!}{
		\begin{tabular}{l|c|ccccc}
			\toprule
			Methods & Resolution & PSNR $\uparrow$ & SSIM $\uparrow$ & LPIPS $\downarrow$ & FID $\downarrow$ \\
			\midrule
			LIME & 3840$\times$2160 & 17.361 & 0.499 & 0.455 & 52.494 \\
			DRBN & 3840$\times$2160 & 16.517 & 0.661 & 0.456 & 83.900 \\
			Zero-DCE & 3840$\times$2160 & 17.088 & 0.629 & 0.510 & 54.251 \\
			MIRNet & 1280$\times$720 & 19.395 & 0.770 & 0.403 & 62.390 \\
			EnlightenGAN & 3840$\times$2160 & 17.391 & 0.666 & 0.473 & 64.788 \\
			RUAS & 3840$\times$2160 & 11.765 & 0.647 & 0.491 & 89.438 \\
			DDIM$^{\ast}$ & 960$\times$540 & 16.820 & 0.699 & 0.445 & 59.573 \\
			SCI & 3840$\times$2160 & 15.467 & 0.586 & 0.524 & 44.093 \\
			URetinex-Net & 2560$\times$1440 & \underline{20.966} & 0.762 & \underline{0.368} & \underline{38.780} \\
			SNRNet & 1920$\times$1080 & 15.967 & 0.730 & 0.424 & 60.009 \\
			Uformer$^{\ast}$ & 960$\times$540 & 18.112 & \underline{0.793} & 0.421 & 75.229 \\
			Restormer$^{\ast}$ & 960$\times$540 & 18.072 & 0.760 & 0.446 & 53.972 \\
			Palette$^{\ast}$ & 720$\times$480 & 9.973 & 0.540 & 0.729 & 86.313 \\
			UHDFour$_{2\times}$ & 1920$\times$1080 & 11.958 & 0.459 & 0.530 & 128.927 \\
			WeatherDiff$^{\ast}$ & 1920$\times$1080 & 16.693 & 0.769 & 0.403 & 102.380 \\
			DiffLL(Ours) & 2560$\times$1440 & \textbf{21.356} & \textbf{0.803} & \textbf{0.356} & \textbf{32.686} \\
			\midrule
			UHDFour$_{8\times}$ & 3840$\times$2160 & \textbf{26.226} & \textbf{0.872} & \textbf{0.239} & \textbf{20.851} \\
			\bottomrule
	\end{tabular}}
	\label{tab:quantitative_UHD}
\end{table}

To validate that, we compare the proposed method with competitive methods that perform well in small-resolution paired datasets on the UHD-LL~\cite{UHD_ICLR} test set, including LIME, DRBN, Zero-DCE, MIRNet, EnlightenGAN, RUAS, SCI, URetinex-Net, SNRNet, Uformer, Restormer, UHDFour, DDIM, Palette, WeatherDiff, and GDP. As mentioned in Table~\ref{tab:inference_time}, most methods are unable to handle large-resolution images, we, therefore, follow~\cite{UHD_ICLR} to downsample the input to the maximum size that the models can handle and then resize the results to the original resolution for performance evaluation. As shown in Table~\ref{tab:quantitative_UHD}, our method achieves state-of-the-art quantitative performance compared with all comparison methods trained on the LOLv1 training set. In terms of distortion metrics, our method outperforms the second-best models, URetinex-Net and Uformer, by 0.390dB(=21.356-20.966) and 0.010(=0.803-0.793) in PSNR and SSIM, respectively. For perceptual metrics, i.e., LPIPS and FID, our method yields the lowest perceptual scores among all comparison methods, indicating our ability to generate images with satisfactory visual quality. Even when compared with UHDFour$_{8\times}$, which is specifically designed for the UHD LLIE task and trained on the UHD-LL training set, our method does not present significant inferiority. These findings demonstrate the generalization ability of our method for high-resolution low-light image restoration.
\begin{table*}[!t]
	\centering
	\caption{Quantitative comparison of different methods on DICM~\cite{DICM}, MEF~\cite{MEF}, LIME~\cite{LIME}, NPE~\cite{NPE}, and VV datasets. The best results are highlighted in \textbf{bold} and the second best results are \underline{underlined}. $\ast$ denotes the methods are retrained on the LOLv1~\cite{RetinexNet} training set and BRI. denotes BRISQUE~\cite{BRISQUE}.}
	\resizebox{\linewidth}{!}{
		\begin{tabular}{l|ccc|ccc|ccc|ccc|ccc|ccc}
			\toprule
			\multirow{2}[4]{*}{Methods} & \multicolumn{3}{c|}{DICM} & \multicolumn{3}{c|}{MEF} & \multicolumn{3}{c|}{LIME} & \multicolumn{3}{c|}{NPE} & \multicolumn{3}{c|}{VV} & \multicolumn{3}{c}{AVG.} \\
			\cmidrule{2-19}      & NIQE $\downarrow$ & BRI. $\downarrow$ & PI $\downarrow$ & NIQE $\downarrow$ & BRI. $\downarrow$ & PI $\downarrow$ & NIQE $\downarrow$ & BRI. $\downarrow$ & PI $\downarrow$ & NIQE $\downarrow$ & BRI. $\downarrow$ & PI $\downarrow$ & NIQE $\downarrow$ & BRI. $\downarrow$ & PI $\downarrow$ & NIQE $\downarrow$ & BRI. $\downarrow$ & PI $\downarrow$ \\
			\midrule
			LIME & 4.476 & 27.375 & 4.216 & 4.744 & 39.095 & 5.160 & 5.045 & 32.842 & 4.859 & 4.170 & 28.944 & 3.789 & 3.713 & 18.929 & 3.335 & 4.429 & 29.437 & 4.272 \\
			DRBN & 4.369 & 30.708 & 3.800 & 4.869 & 44.669 & 4.711 & 4.562 & 34.564 & 3.973 & 3.921 & 25.336 & 3.267 & 3.671 & 24.945 & 3.117 & 4.278 & 32.045 & 3.774 \\
			Zero-DCE & 3.951 & 23.350 & 3.149 & \underline{3.500} & 29.359 & \textbf{2.989} & 4.379 & 26.054 & \underline{3.239} & 3.826 & 21.835 & 2.918 & 5.080 & 21.835 & 3.307 & 4.147 & 24.487 & 3.120 \\
			MIRNet & 4.021 & 22.104 & 3.691 & 4.202 & 34.499 & 3.756 & 4.378 & 28.623 & 3.398 & 3.810 & 23.658 & 3.205 & 3.548 & 22.897 & 2.911 & 3.992 & 26.356 & 3.392 \\
			EnlightenGAN & 3.832 & 19.129 & 3.256 & 3.556 & 26.799 & 3.270 & \underline{4.249} & 22.664 & 3.381 & 3.775 & 21.157 & 2.953 & 3.689 & \textbf{14.153} & 2.749 & 3.820 & 20.780 & 3.122 \\
			RUAS & 7.306 & 46.882 & 5.700 & 5.435 & 42.120 & 4.921 & 5.322 & 34.880 & 4.581 & 7.198 & 48.976 & 5.651 & 4.987 & 35.882 & 4.329 & 6.050 & 41.748 & 5.036 \\ 
			DDIM$^{\ast}$ & 3.899 & 19.787 & 3.213 & 3.621 & 28.614 & 3.376 & 4.399 & 24.474 & 3.459 & 3.679 & 18.996 & 2.870 & 3.640 & 16.879 & 2.669 & 3.848 & 21.750 & 3.118 \\
			SCI & 4.519 & 27.922 & 3.700 & 3.608 & 26.716 & 3.286 & 4.463 & 25.170 & 3.376 & 4.124 & 28.887 & 3.534 & 5.312 & 22.800 & 3.648 & 4.405 & 26.299 & 3.509 \\
			URetinex-Net & 4.774 & 24.544 & 3.565 & 4.231 & 34.720 & 3.665 & 4.694 & 29.022 & 3.713 & 4.028 & 26.094 & 3.153 & 3.851 & 22.457 & 2.891 & 4.316 & 27.368 & 3.397 \\
			SNRNet & \underline{3.804} & 19.459 & 3.285 & 4.063 & 28.331 & 3.753 & 4.597 & 29.023 & 3.677 & 3.940 & 28.419 & 3.278 & 3.761 & 23.672 & 2.903 & 4.033 & 25.781 & 3.379 \\
			Uformer$^{\ast}$ & 3.847 & 19.657 & 3.180 & 3.935 & \underline{25.240} & 3.582 & 4.300 & 21.874 & 3.565 & \underline{3.510} & 16.239 & 2.871 & \textbf{3.274} & 19.491 & \underline{2.618} & \underline{3.773} & 20.500 & 3.163 \\
			Restormer$^{\ast}$ & 3.964 & 19.474 & 3.152 & 3.815 & 25.322 & 3.436 & 4.365 & 22.931 & 3.292 & 3.729 & 16.668 & \underline{2.862} & 3.795 & 16.974 & 2.712 & 3.934 & 20.274 & 3.091 \\
			Palette$^{\ast}$ & 4.118 & \underline{18.732} & 3.425 & 4.459 & 25.602 & 4.205 & 4.485 & 20.551 & 3.579 & 3.777 & 16.006 & 3.018 & 3.847 & 16.106 & 2.986 & 4.137 & \underline{19.399} & 3.443 \\
			UHDFour$_{2\times}$ & 4.575 & 26.926 & 3.684 & 4.231 & 29.538 & 4.124 & 4.430 & \underline{20.263} & 3.813 & 4.049 & \underline{15.934} & 3.135 & 3.867 & 15.297 & 2.894 & 4.230 & 21.592 & 3.530 \\
			WeatherDiff$^{\ast}$ & \textbf{3.773} & 20.387 & \underline{3.130} & 3.753 & 30.480 & 3.312 & 4.312 & 28.090 & 3.424 & 3.677 & 20.262 & 2.878 & \underline{3.472} & 18.070 & 2.656 & 3.797 & 23.458 & \underline{3.080} \\
			GDP & 4.358 & 19.294 & 3.552 & 4.609 & 34.859 & 4.115 & 4.891 & 27.460 & 3.694 & 4.032 & 19.527 & 3.097 & 4.683 & 20.910 & 3.431 & 4.514 & 24.410 & 3.578 \\
			DiffLL(Ours) & 3.806 & \textbf{18.584} & \textbf{3.011} & \textbf{3.427} & \textbf{24.165} & \underline{3.011} & \textbf{3.777} & \textbf{19.843} & \textbf{3.074} & \textbf{3.427} & \textbf{15.789} & \textbf{2.597} & 3.507 & \underline{14.644} & \textbf{2.562} & \textbf{3.589} & \textbf{18.605} & \textbf{2.851} \\
			\bottomrule
	\end{tabular}}
	\label{tab:evaluation_unpaired}
\end{table*}
\begin{figure*}[!t]
	\centering
	\includegraphics[width=\linewidth]{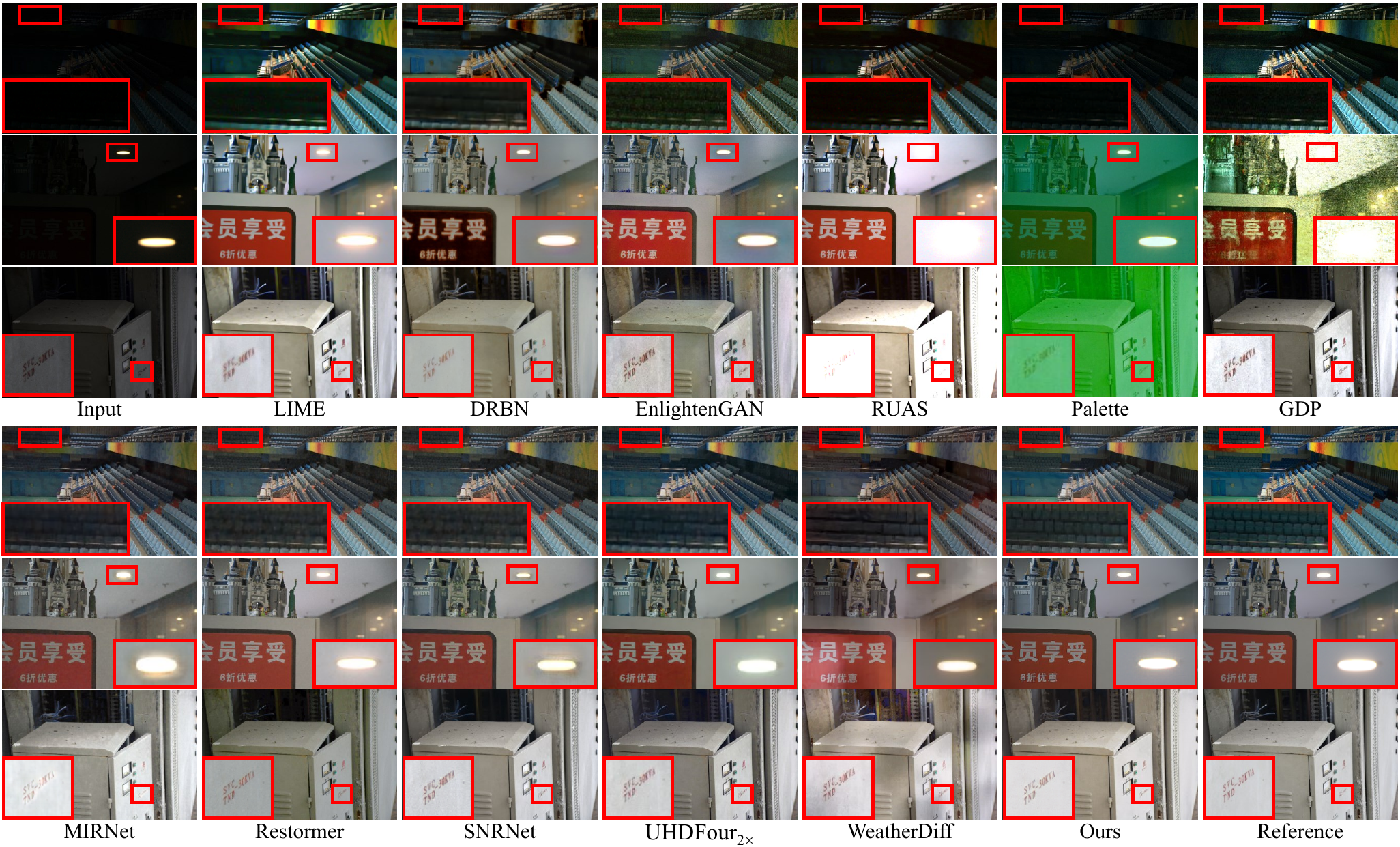}
	\caption{Qualitative comparison of our method and competitive methods on the LOLv1~\cite{RetinexNet} (row 1), LOLv2-real~\cite{LOLV2} (row 2), and LSRW~\cite{R2RNet} (row 3) test sets. Error-prone regions are highlighted with red boxes, best viewed by zooming in.}
	\label{fig:visual_compare_paired}
\end{figure*}
\begin{figure*}[!t]
	\centering
	\includegraphics[width=\linewidth]{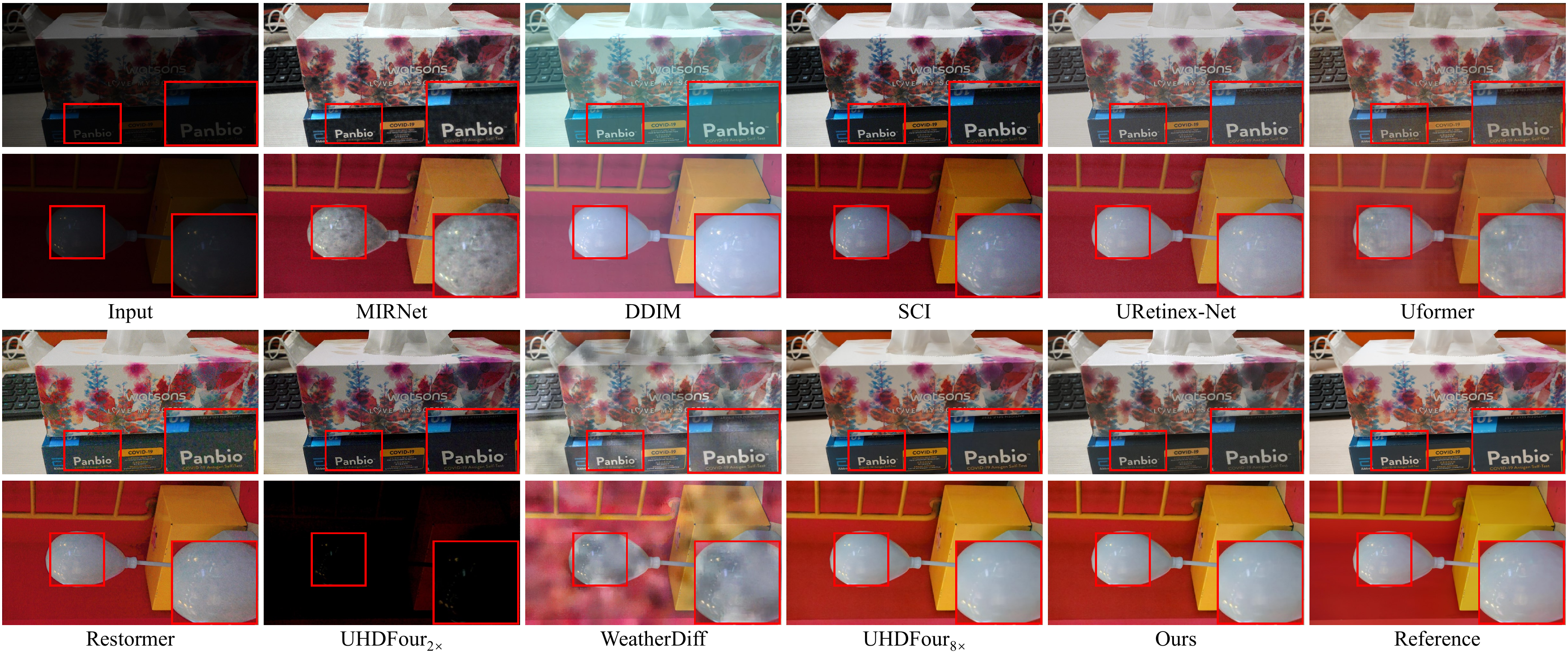}
	\caption{Qualitative comparison of our method and competitive methods on the UHD-LL~\cite{UHD_ICLR} test set. Error-prone regions are highlighted with red boxes, best viewed by zooming in. Note that UHDFour$_{8\times}$ is trained on the UHD-LL training set, while other comparison supervised methods and our method are trained on the simpler LOLv1~\cite{RetinexNet} training set.}
	\label{fig:visual_UHD}
\end{figure*}
\begin{figure*}[!t]
	\centering
	\includegraphics[width=\linewidth]{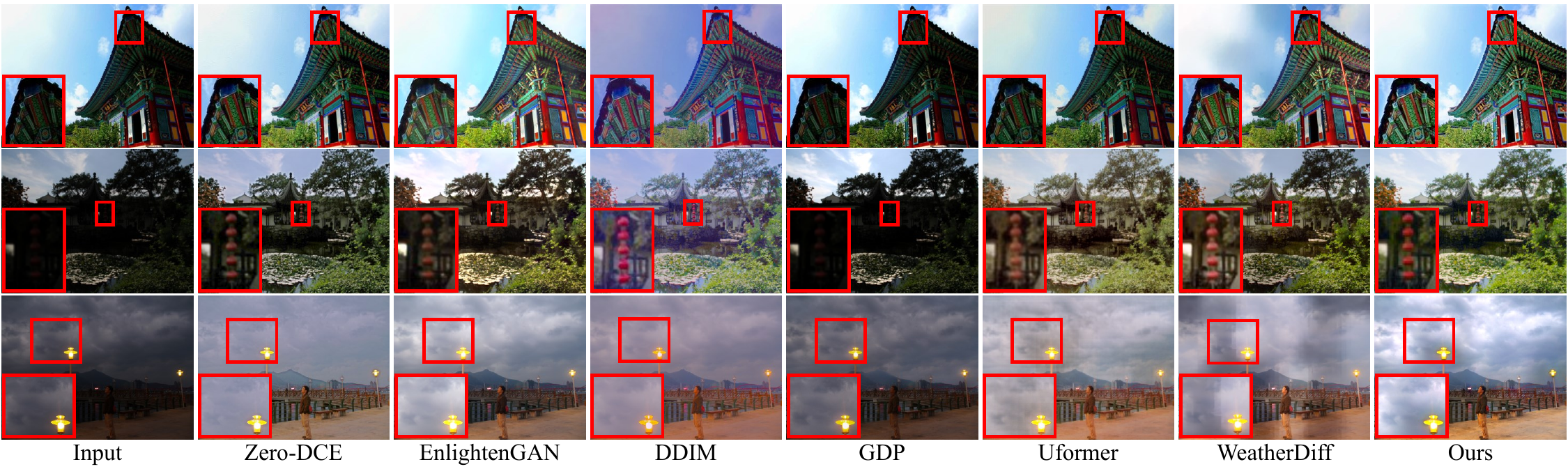}
	\caption{Qualitative comparison of our method and competitive methods on the DICM~\cite{DICM} (row 1), MEF~\cite{MEF} (row 2), and NPE~\cite{NPE} (row 3) datasets. Error-prone regions are highlighted with red boxes, best viewed by zooming in.}
	\label{fig:visual_compare_unpaired}
\end{figure*}

Moreover, we also compared the proposed DiffLL with competitive methods on five unpaired datasets to further validate the effectiveness of our method. Three non-reference perceptual metrics NIQE~\cite{NIQE}, BRISQUE~\cite{BRISQUE}, and PI~\cite{PI} are utilized to evaluate the visual quality of the enhanced results, the lower the metrics the better the visual quality. As shown in Table~\ref{tab:evaluation_unpaired}, our method achieves the lowest NIQE score on the MEF, LIME, and NPE datasets, and is comparable to the best methods on the DICM and VV datasets. For BRISQUE and PI scores, we achieve the best results on four datasets as well as the second-best results on the remaining dataset. In summary, our method achieves the lowest average scores on all three metrics, which proves that our method has a better generalization ability to unseen real-world scenes.

\textbf{Qualitative Comparison.}
We present qualitative comparisons of our method and state-of-the-art methods on the paired datasets in Fig.~\ref{fig:visual_compare_paired}. The images in rows 1-3 are selected from LOLv1, LOLv2-real, and LSRW test sets, respectively. The results presented in rows 1 and 3 show that DRBN, MIRNet, RUAS, and GDP are unable to properly improve the contrast of the images, yielding underexposed or overexposed results. Likewise, EnlightenGAN suffers from noise amplification, while LIME, SNRNet, Restormer, UHDFour$_{2\times}$ and WeatherDiff produce over-smoothed textures and artifacts. Moreover, as depicted in row 2, the results restored by Palette and GDP are concomitant with chaotic content and color distortion, other methods typically introduce artifacts around the lamp, whereas WeatherDiff avoids this distortion in that particular area but exhibits artifacts elsewhere. In contrast, our method effectively improves global and local contrast, reconstructs sharper details, and suppresses noise, resulting in visually satisfactory results in all cases. The visual comparisons on the UHD-LL~\cite{UHD_ICLR} test set is illustrated in Fig.~\ref{fig:visual_UHD}. As we can see that previous methods appear incorrect exposure, color distortion, noise amplification, or artifacts, thereby undermining the overall visual quality. In contrast, our method effectively improves global contrast and presents a vivid color without introducing chaotic content.
\begin{figure*}[!t]
	\centering
	\includegraphics[width=\linewidth]{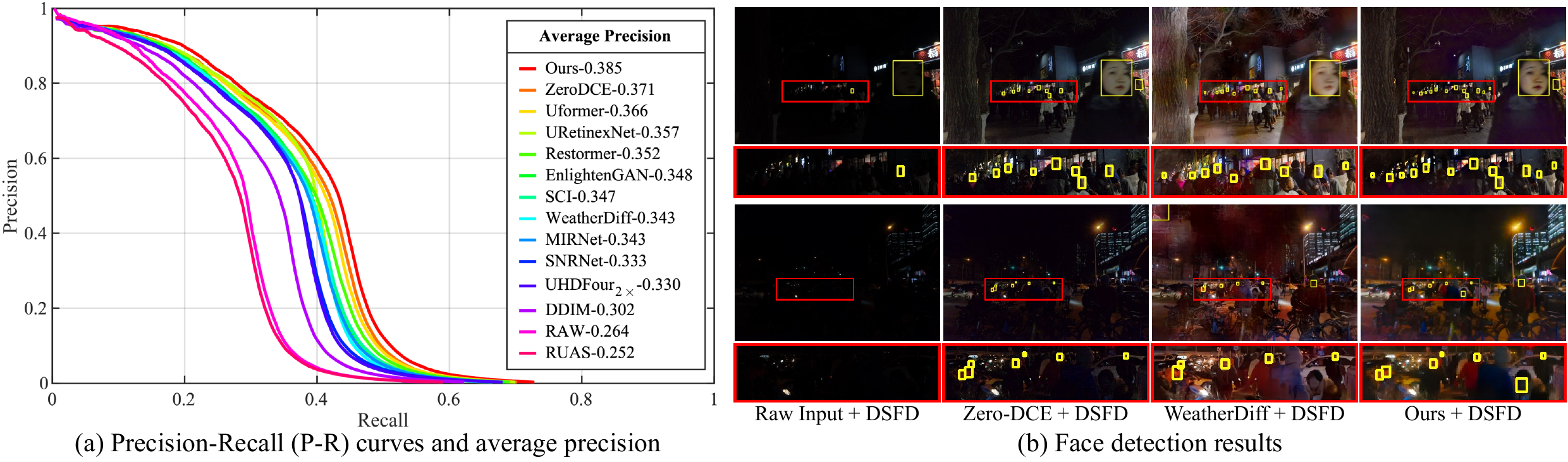}
	\caption{Comparison of face detection results before and after enhanced by different methods on the DARK FACE dataset~\cite{DarkFace}.}
	\label{fig:face_detection}
\end{figure*}

We also provide visual comparisons of our method and competitive methods on the unpaired datasets in Fig.~\ref{fig:visual_compare_unpaired}, where the images in rows 1-3 are selected from DICM, MEF, and NPE datasets, respectively. The previous learning-based LLIE methods and GDP fail to relight the building and the lantern in rows 1 and 2 compared to our method and WeatherDiff, but WeatherDiff produces undesirable content in the sky. In row 3, Zero-DCE, EnlightenGAN, and DDIM smooth the cloud texture, as Uformer and WeatherDiff produce gridding effects and messy content. Our method successfully restores the illumination of the image and reconstructs the details, which demonstrates that our method can generalize well to unseen scenes. For more qualitative results, please refer to the supplementary material.

\subsection{Low-Light Face Detection}\label{subsec:Low-Light Face Detection}
In this section, we investigate the impact of LLIE methods as pre-processing on improving the face detection task under weakly illuminated conditions. Specifically, we take the DARK FACE dataset~\cite{DarkFace} consisting of 10,000 images captured in real-world nighttime scenes for comparison, of which 6,000 images are selected as the training set, and the rest are adopted for testing. Since the bounding box labels of the test set are not publicly available, we utilize the training set for evaluation, applying our method and comparison methods as a pre-processing step, followed by the well-known learning-based face detector DSFD~\cite{DSFD}. We illustrate the precision-recall (P-R) curves in Fig.~\ref{fig:face_detection}(a) and compare the average precision (AP) under the IoU threshold of 0.5 using the official evaluation tool. As we can see that our proposed method achieves superior performance in the high recall area, and utilizing our approach as a pre-processing step leads to AP improvement from 26.4\% to 38.5\%. Observing the visual examples in Fig.~\ref{fig:face_detection}(b), our method relights the face regions while maintaining well-illuminated regions, enabling DSFD more robust in weakly illuminated scenes.

\subsection{Ablation Study}\label{subsec:Ablation Study}
In this section, we conduct a set of ablation studies to measure the impact of different component configurations employed in our method. We employ the implementation details described in Sec.~\ref{subsec:Experimental Settings} to train all models and subsequently evaluate their performances on the LOLv1~\cite{RetinexNet} test set for analysis. Detailed experiment settings are discussed below.

\textbf{Wavelet Transformation Scale.} We first validate the impact of performing diffusion processes on the average coefficient at different wavelet scales $K$. As reported in Table~\ref{tab:efficiency_ablation}, when we apply the WCDM to restore the average coefficient $A_{low}^{1}$, i.e., $K=1$, leads to overall performance improvements while resulting in the inference time increase. As the wavelet scale increases, the spatial dimension of the average coefficient decreases by $4^{K}$ relative to the original image, which causes performance degradation due to the information richness reduction but lowers the inference time. For a trade-off between performance and efficiency, we choose $K=2$ as the default setting. From another perspective, even if we perform three times wavelet transformation, i.e., $K=3$, our method is also comparable to the state-of-the-art methods reported in Table~\ref{tab:evaluation_paired}, which demonstrates that our proposed wavelet-based diffusion model can be applied to restore higher-resolution images.
\begin{table}[!t]
	\centering
	\caption{Ablation studies of various settings on the wavelet transformation scale and sampling step, please refer to the text for more details. The results using default settings are \underline{underlined}.}
	\resizebox{\linewidth}{!}{
		\begin{tabular}{c|c|ccccc}
			\toprule
			Wavelet scale & Sampling step  & PSNR $\uparrow$ & SSIM $\uparrow$ & LPIPS $\downarrow$ & FID $\downarrow$ & Time (s) $\downarrow$ \\
			\midrule
			\multirow{4}[2]{*}{$K=1$} & $S=5$ & 26.256 & 0.855 & 0.188 & 40.884 & 0.198 \\
			& $S=10$ & 26.444 & 0.858 & 0.182 & 40.143 & 0.380 \\
			& $S=20$ & 26.385 & 0.856 & 0.179 & 40.001 & 0.757 \\
			& $S=30$ & 26.211 & 0.850 & 0.190 & 41.023 & 1.256 \\
			\midrule
			\multirow{4}[2]{*}{$K=2$} & $S=5$ & 25.855 & 0.835 & 0.210 & 49.371 & 0.085 \\
			& $S=10$ & \underline{26.336} & \underline{0.845} & \underline{0.217} & \underline{48.114} & \underline{0.157} \\
			& $S=20$ & 26.246 & 0.844 & 0.212 & 48.071 & 0.285 \\
			& $S=30$ & 26.094 & 0.851 & 0.213 & 49.993 & 0.477 \\
			\midrule
			\multirow{4}[2]{*}{$K=3$} & $S=5$ & 24.408 & 0.817 & 0.283 & 61.234 & 0.066 \\
			& $S=10$ & 25.091 & 0.828 & 0.265 & 57.190 & 0.114 \\
			& $S=20$ & 25.103 & 0.826 & 0.271 & 57.227 & 0.237 \\
			& $S=30$ & 24.796 & 0.822 & 0.279 & 59.326 & 0.401 \\
			\bottomrule
	\end{tabular}}
	\label{tab:efficiency_ablation}
\end{table}

\textbf{Sampling Step.} Another factor that enables our method to be more computational efficiency is that a smaller sampling step $S$ is adopted. We present the performance of our method using different $S$ from 5 to 30 in Table~\ref{tab:efficiency_ablation}. Benefiting from we conduct the denoising process during training, the model learns how to perform denoising with different steps, making variations in sampling steps, whether smaller or larger, have no major impact on the performance of our proposed WCDM. A large sampling step contributes to generating images with better visual quality for diffusion-based image generation methods~\cite{ddpm,ddim}, while for us it only increases the inference time, which indicates that larger sampling steps, e.g., $S=1000$ in GDP~\cite{GDP} and Palette~\cite{palette}, may not be essential for diffusion-based LLIE and other relative image restoration tasks using our proposed framework.

\textbf{High-frequency Restoration Module.}
To validate the effectiveness of our designed high-frequency restoration module (HFRM), we form a baseline by using the proposed wavelet-based conditional diffusion model (WCDM) only to restore the average coefficient. As shown in rows 1-3 of Table~\ref{tab:module_ablation}, the HFRMs utilized to reconstruct the high-frequency coefficients at wavelet scales $k=1$ and $k=2$ improve the overall performance, achieving 2.142dB and 1.440dB gains in terms of PSNR, as well as 0.064 and 0.044 in terms of SSIM, respectively. Since the high-frequency coefficients at $k=1$ have richer details than the coefficients at $k=2$, the HFRM$_{1}$ is thus capable of delivering greater gains. Furthermore, we have formed two HFRM versions dubbed HFRM-v2 and HFRM-v3 to explore the effect of high-frequency coefficients complementarity. Specifically, the HFRM-v2 does not leverage the information in vertical and horizontal directions to complement the details present in the diagonal by using convolutional layers with equal parameters to replace cross-attention layers. On the other hand, the HFRM-v3 utilizes information from the diagonal direction to reinforce the vertical and horizontal details. As reported in rows 4-6 of Table~\ref{tab:module_ablation}, our default HFRM is superior to the other two versions, which demonstrates the effectiveness of high-frequency complementarity and that complementing diagonal details with information in vertical and horizontal directions being more useful than the reverse.
\begin{table}[!t]
	\centering
	\caption{Ablation studies of the effectiveness of our high-frequency restoration module. The results using default settings are \underline{underlined}.}
	\resizebox{0.75\linewidth}{!}{
		\begin{tabular}{l|cccc}
			\toprule
			& PSNR $\uparrow$ & SSIM $\uparrow$ & LPIPS $\downarrow$ & FID $\downarrow$ \\
			\midrule
			WCDM & 21.975 & 0.729 & 0.310 & 98.104 \\
			WCDM + HFRM$_{2}$  & 23.415 & 0.773 & 0.259 & 76.193 \\
			WCDM + HFRM$_{1}$ & 24.117 & 0.793 & 0.251 & 62.210 \\
			\midrule
			WCDM + HFRM-v2 & 24.145 & 0.822 & 0.266 & 67.909 \\
			WCDM + HFRM-v3 & 25.630 & 0.824 & 0.221 & 51.466 \\
			\midrule
			Default & \underline{26.336} & \underline{0.845} & \underline{0.217} & \underline{48.114} \\
			\bottomrule
	\end{tabular}}
	\label{tab:module_ablation}
\end{table}
\begin{table}[!ht]
	\centering
	\caption{Ablation studies of the loss function terms. The results using default settings are \underline{underlined}. `w/o' denotes without.}
	\resizebox{0.8\linewidth}{!}{
		\begin{tabular}{l|cccc}
			\toprule
			& PSNR $\uparrow$ & SSIM $\uparrow$ & LPIPS $\downarrow$ & FID $\downarrow$ \\
			\midrule
			w/o $\mathcal{L}_{diff}$ & 24.238 & 0.823 & 0.314 & 89.318 \\
			w/o $\mathcal{L}_{content}$ & 23.821 & 0.808 & 0.283 & 67.502 \\
			w/o $\mathcal{L}_{detail}$ & 25.413 & 0.835 & 0.249 & 57.663 \\
			\midrule
			vanilla $\mathcal{L}_{diff}$ (Eq.\ref{eq:8}) & 25.956 & 0.838 & 0.221 & 50.467 \\
			\midrule
			Default & \underline{26.306} & \underline{0.845} & \underline{0.217} & \underline{48.114} \\
			\bottomrule
	\end{tabular}}
	\label{tab:loss_ablation}
\end{table}

\textbf{Loss Function.}
To validate the effectiveness of the proposed loss functions, we conduct experiments by individually removing each component from the default setting, where the quantitative results are reported in Table~\ref{tab:loss_ablation}. As shown in row 1, the removal of the diffusion loss $\mathcal{L}_{diff}$ results in decreases in all perceptual metrics, as the generative capacity of diffusion models relies heavily on this component. The incorporation of content loss $\mathcal{L}_{content}$ yields noticeable improvements, especially the distortion metrics, which can be improved by 2.485dB and 0.037 for PSNR and SSIM, respectively, as shown in row 2 and row 5. The detail preserve loss $\mathcal{L}_{detail}$ is designed to reconstruct more image details, thus its removal causes performance degradation. However, such degradation is not significant relative to removing the content loss, which reveals the importance of the content loss. From another perspective, the content loss should be integrated with our proposed training strategy, which also illustrates its effectiveness. Furthermore, as described in Sec.~\ref{subsec:Wavelet-based Conditional Diffusion Models}, we incorporate an extra loss term into the vanilla diffusion loss, i.e., Eq.\ref{eq:8}, to promote the restoration of the average coefficient. To validate the effectiveness of the auxiliary term, we employ vanilla diffusion loss to optimize the noise estimator network. As shown in row 4, the inclusion of the auxiliary term proves to be beneficial in enhancing the visual fidelity of the restored images, thereby leading to the overall performance improved.
\begin{table}[!t]
	\centering
	\caption{Ablation studies of the training strategy of our wavelet-based conditional diffusion model. The results using default settings are \underline{underlined}.}
	\resizebox{\linewidth}{!}{
		\begin{tabular}{l|cccc}
			\toprule
			Methods & PSNR $\uparrow$ & SSIM $\uparrow$ & LPIPS $\downarrow$ & FID $\downarrow$ \\
			\midrule
			DDIM$^{\ast}$ & 16.521$_{\pm 0.544}$ & 0.776$_{\pm 0.002}$ & 0.376$_{\pm 0.001}$ & 84.071${_\pm 32.909}$ \\
			Palette$^{\ast}$ & 11.771$_{\pm 0.254}$ & 0.561$_{\pm 0.001}$ & 0.498$_{\pm 0.001}$ & 108.291${_\pm 8.019}$ \\
			WeatherDiff$^{\ast}$ & 17.913$_{\pm 0.182}$ & 0.811$_{\pm 0.001}$ & 0.272$_{\pm 0.001}$ & 73.903${_\pm 5.265}$\\
			\midrule
			Ours$_{\operatorname{FD}}$ & 18.125$_{\pm 0.212}$ & 0.789$_{\pm 0.001}$ & 0.273$_{\pm 0.002}$ & 64.676$_{\pm 5.544}$ \\
			Ours$_{\operatorname{FD+DP}}$ & \underline{26.336$_{\pm 1\times 10^{-5}}$} & \underline{0.845$_{\pm 2\times 10^{-7}}$} & \underline{0.217$_{\pm 6\times 10^{-9}}$} & \underline{48.114$_{\pm 0.006}$} \\
			\bottomrule
	\end{tabular}}
	\label{tab:training_ablation}
\end{table}
\begin{figure}[!t]
	\centering
	\includegraphics[width=\linewidth]{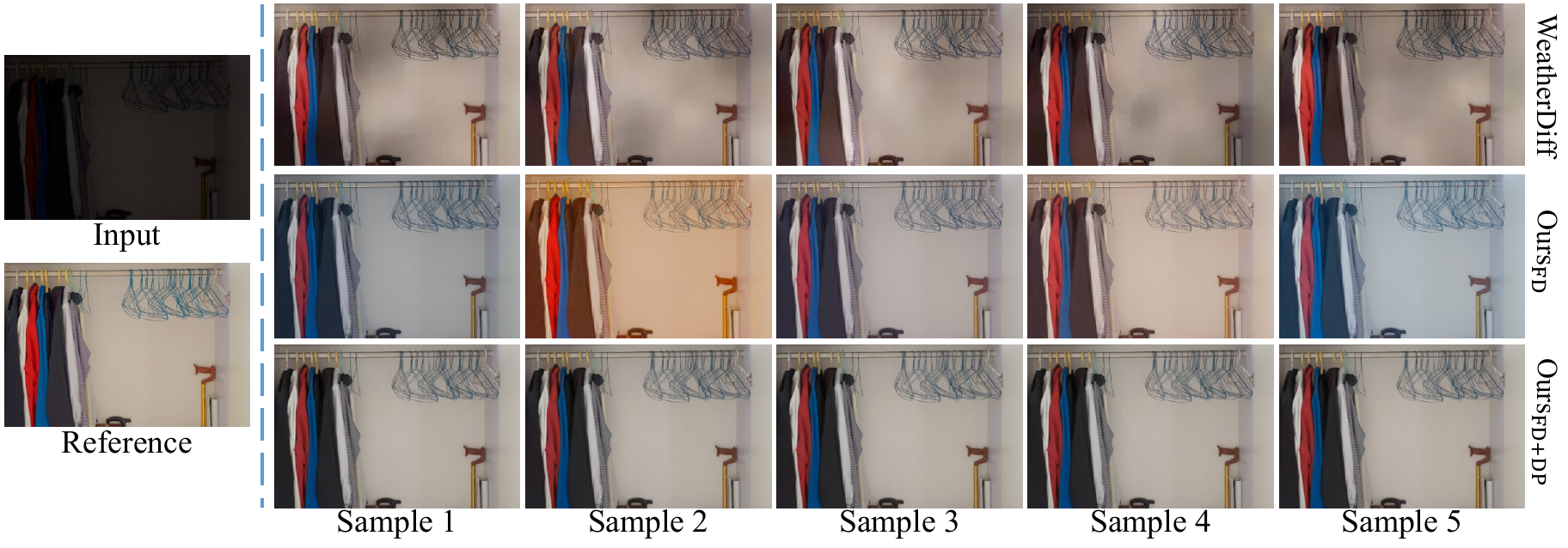}
	\caption{The sampling results from five evaluations, in which performing forward diffusion only in the training phase results in diverse outputs, while our training strategy contributes to generating results with consistency.}
	\label{fig:diversity}
\end{figure}

\textbf{Training Strategy.}
The content diversity caused by randomly sampled noise during the denoising process of diffusion models is undesirable for some image restoration tasks, such as LLIE, dehazing, deraining, etc. As described in Sec.~\ref{subsec:Wavelet-based Conditional Diffusion Models}, we address this issue by incorporating the denoising process (DP) in both training and inference phases to achieve consistency. As shown in Table~\ref{tab:training_ablation}, we report the means and variances of five evaluations for three diffusion-based methods, our method that performs forward diffusion (FD) only in the training phase denoted as Ours$_{\operatorname{FD}}$, and our method with the default training strategy denoted as Ours$_{\operatorname{FD+DP}}$. By incorporating DP into the training phase, our method not only achieves the best performance but also aids in circumventing content diversity and yields minimal variance, which proves the superiority of our training strategy. The restored results of five evaluations illustrated in Fig.~\ref{fig:diversity} show that performing forward diffusion only in the training phase leads to diverse results, while our default training strategy not only mitigates the appearance of chaotic content and color distortion but also facilitates generating images with consistency.

\subsection{Limitations}
Although our method is effective in restoring images captured in weakly illuminated conditions, it does not work as well in extremely low-light environments as images captured in such conditions suffer from greater information loss and are more challenging to restore. In addition, some real-time methods can be applied directly to the low-light video enhancement task, whereas our method is not yet efficient enough to be applied directly to this task. Another natural limitation of our model is its inherently limited capacity to only generalize to the LLIE task observed at training time, and investigating the effectiveness of the proposed approach for other image restoration tasks will be our future work.

\section{Conclusions}\label{sec:conclusion}
We have presented DiffLL, a diffusion-based framework for robust and efficient low-light image enhancement. Technically, we propose a wavelet-based conditional diffusion model that leverages the generative ability of diffusion models and wavelet transformation to produce visually satisfactory results while reducing inference time and computational resource consumption. Moreover, to circumvent content diversity during inference, we perform the denoising process in both the training and inference phases, enabling the model to learn stable sampling and obtain restored results with consistency. Additionally, we design a high-frequency restoration module to complement the diagonal details with vertical and horizontal information for better fine-grained details reconstruction. Experimental results on publicly available benchmarks demonstrate that our method outperforms competitors both quantitatively and qualitatively while being computationally efficient. The results of the low-light face detection also show the practical values of our method.

\section*{Acknowledgements}
This work was supported by National Natural Science Foundation of China (NSFC) under grant No.62372091 and Sichuan Science and Technology Program of China under grant No.2023NSFSC0462.

% Bibliography
\bibliographystyle{ACM-Reference-Format}
\bibliography{sample-bibliography}
\end{sloppypar}
\end{document}